\documentclass{article}

\PassOptionsToPackage{numbers, compress}{natbib}
\PassOptionsToPackage{prologue,dvipsnames}{xcolor}

\usepackage[preprint]{neurips_2025}

\usepackage[utf8]{inputenc} 
\usepackage[T1]{fontenc}    
\usepackage{hyperref}       
\usepackage{url}            
\usepackage{booktabs}       
\usepackage{amsfonts}       
\usepackage{nicefrac}       
\usepackage{microtype}      
\usepackage{xcolor}         
\usepackage{graphicx}
\usepackage{amsmath}
\usepackage{float}
\usepackage{subcaption}
\usepackage{caption}
\usepackage{tcolorbox}

\definecolor{midnightgreen}{rgb}{0.0, 0.29, 0.33}

\newcommand{\method}{\mbox{\textsc{AutoRule}}}

\title{\method: Reasoning Chain-of-thought Extracted Rule-based Rewards Improve Preference Learning}

\author{%
  Tevin Wang \quad Chenyan Xiong\\
  School of Computer Science\\
  Carnegie Mellon University\\
  \texttt{\{tevinw, cx\}@andrew.cmu.edu} \\
}

\begin{document}

\maketitle

\begin{abstract}
Rule-based rewards offer a promising strategy for improving reinforcement learning from human feedback (RLHF), but current approaches often rely on manual rule engineering. We present \method{}, a fully automated method for extracting rules from preference feedback and formulating them into rule-based rewards. \method{} extraction operates in three stages: it leverages a reasoning model to interpret user preferences, identifies candidate rules from the reasoning chain of these interpretations, and synthesizes them into a unified rule set. Leveraging the finalized rule set, we employ language-model verifiers to compute the fraction of rules satisfied by each output, using this metric as an auxiliary reward alongside the learned reward model during policy optimization. Training a Llama-3-8B model with \method{} results in a 28.6\% relative improvement in length-controlled win rate on AlpacaEval2.0, and a 6.1\% relative gain in second-turn performance on a held-out MT-Bench subset, compared to a GRPO baseline trained with the same learned reward model but without the rule-based auxiliary reward. Our analysis confirms that the extracted rules exhibit good agreement with dataset preference. We find that \method{} demonstrates reduced reward hacking compared to a learned reward model when run over two episodes. Finally, our case study suggests that the extracted rules capture unique qualities valued in different datasets. The extracted rules are provided in the appendix, and the code is open-sourced at \url{https://github.com/cxcscmu/AutoRule}.
\end{abstract}
\section{Introduction}

Reinforcement learning from human feedback (RLHF) has become a cornerstone technique for aligning large language models (LLMs) with human values and enhancing their ability to follow human instructions \cite{NEURIPS2022_b1efde53}. RLHF and related preference-based optimization approaches have been utilized in top industry models like GPT-4 \cite{openai2024gpt4technicalreport}, Gemini \cite{geminiteam2025geminifamilyhighlycapable}, Claude \cite{claude3} and Llama 3 \cite{grattafiori2024llama3herdmodels}. RL-based post-training methodologies have also been leveraged to enhance the reasoning capabilities of LLMs. Notably, a key advancement in the release of  Deepseek-R1 is the adoption of rule-based rewards for accuracy and formatting in place of neural rewards, as a strategy to mitigate reward hacking \cite{deepseekai2025deepseekr1incentivizingreasoningcapability}. Rule-based rewards for reasoning tasks are particularly effective because they provide objective, verifiable criteria that govern policy behavior. When a language model's output satisfies these rules, it can be reliably considered an accurate response.

While rule-based rewards work well for reasoning tasks, utilizing them for preference alignment in language models remains challenging. Unlike domains such as code or mathematics, where explicit rule-based verifiers can be constructed, preference alignment is difficult because human preferences are often ambiguous and subjective. Existing industry approaches typically rely on expert-crafted rules \cite{glaese2022improvingalignmentdialogueagents, NEURIPS2024_c4e380fb} or large-scale crowd annotations \cite{bai2022constitutionalaiharmlessnessai}, which can be costly and difficult to scale.

To overcome these limitations, we introduce \method{}, an automatic rule extraction framework that leverages the reasoning capabilities of advanced LLMs to derive alignment rules directly from preference data. Our approach extracts explicit rules from model-generated reasoning chains, moving beyond reliance on hand-crafted or crowd-sourced rules. During RL training, a LLM-as-a-judge \cite{NEURIPS2023_91f18a12} verifier assesses each candidate response for compliance with the extracted rules, and the resulting rule scores are aggregated to form a composite rule-based reward. This reward is then integrated with the standard model reward to guide policy optimization.

To extract rules, \method{} follows the following pipeline. Given a pair of model outputs and an associated preference label, we first prompt a reasoning-capable LLM to generate a step-by-step justification for the preferred output. Next, the LLM is tasked with extracting explicit rule-like statements from its reasoning process. These candidate rules are aggregated across the training set, after which the LLM synthesizes a consolidated rule set. We hypothesize that leveraging the logical structure of reasoning chains enables the extraction of more precise and actionable rules that better capture the underlying preference criteria.

We empirically validate our approach through comprehensive experiments. First, we show that rule-based scores—computed either individually or cumulatively—using Llama 3 8B Instruct \cite{grattafiori2024llama3herdmodels} as the verifier, are in good agreement with preferences on both the UltraFeedback \cite{cui2024ultrafeedback} and MT-Bench Human Judgment \cite{NEURIPS2023_91f18a12} datasets. Next, we post-train the base Llama-3-8B model on UltraFeedback data using the standard RLHF pipeline but replacing traditional PPO with GRPO \cite{shao2024deepseekmathpushinglimitsmathematical} and integrating \method{} as the reward mechanism. We benchmark our method against several baselines, including vanilla PPO and GRPO with model-only rewards, evaluating on UltraFeedback win rates, AlpacaEval 2.0, and MT-Bench. Across all three evaluations, \method{} consistently outperforms the baselines.

Additionally, reward hacking experiments demonstrate the ability of \method{}'s rule-based rewards to mitigate reward model overoptimization. Ablation studies comparing rule extraction from reasoning chains versus justifications support the efficacy of leveraging reasoning chains in \method{}. Furthermore, qualitative analysis reveals that rules derived from UltraFeedback predominantly emphasize conversational quality, while those extracted from MT-Bench prioritize instruction adherence and robustness on more complex tasks. 

In summary, our key contributions are three-fold:
\begin{itemize}
    \item We introduce \method{}, a framework for automatic extraction of alignment rules from preference data via LLM-generated reasoning chains.
    \item We show that rule-based rewards derived via \method{} lead to improved preference alignment and instruction following compared to standard preference optimization baselines.
    \item We demonstrate that \method{} reduces reward hacking and yields interpretable, dataset-adaptive rules.
\end{itemize}
\section{Related Work}
\label{sec:related-work}

Reinforcement learning from human feedback (RLHF) is a standard framework for aligning large language models (LLMs) with human preferences~\citep{NEURIPS2022_b1efde53}. RLHF typically involves: (1) supervised fine-tuning on human-annotated responses; (2) training a reward model to predict human preferences; and (3) reinforcement learning, often via proximal policy optimization (PPO)~\citep{DBLP:journals/corr/SchulmanWDRK17}, using the reward model as the optimization signal. Recent work has explored more efficient approaches to the RL stage, such as Direct Preference Optimization (DPO)~\citep{NEURIPS2023_a85b405e}, which eliminates the reward model, and Group Relative Policy Optimization (GRPO)~\citep{shao2024deepseekmathpushinglimitsmathematical}, which uses relative rewards from groups of outputs.

A well-documented challenge in RLHF with learned reward models is \emph{reward hacking}~\citep{bai2022traininghelpfulharmlessassistant, stiennon2022learningsummarizehumanfeedback, pmlr-v202-gao23h}, in which models exploit idiosyncrasies of the reward model to achieve high reward without genuinely improving response quality. For example, \citet{NEURIPS2024_f25d75fc} find that reward models may overfit to superficial features, such as response length, that do not generalize to the true distribution of human preferences. Supporting this, \citet{singhal2024a} show that optimizing solely for response length during PPO can yield performance comparable to using a learned reward model, indicating that reward models often capture simple heuristics rather than more nuanced aspects of response quality. 

Several strategies have been proposed to mitigate reward hacking, including modifying reward model architectures and adjusting reward scaling. ODIN~\citep{10.5555/3692070.3692382} adds an auxiliary length prediction head to ``disentangle'' length from other features. Reward shaping methods such as PAR~\citep{fu2025rewardshapingmitigatereward} and LSC~\citep{10.5555/3692070.3694167} apply sigmoid or log-sigmoid transformations centered on reference model outputs or percentiles. Other approaches leverage multiple reward models: WARM~\citep{10.5555/3692070.3693780} averages outputs from several reward models to reduce overoptimization, while ArmoRM~\citep{wang-etal-2024-interpretable} combines interpretable reward objectives using a gating mechanism.

A growing strategy for mitigating reward hacking is the adoption of rule-based reward objectives, especially in large-scale industrial LLM deployments. For instance, DeepSeek utilized rule-based rewards during the post-training of DeepSeek-R1~\citep{deepseekai2025deepseekr1incentivizingreasoningcapability}, explicitly prioritizing rule-based criteria over learned reward models to reduce reward hacking. Their approach incorporates two types of rewards: an accuracy reward, which evaluates whether the response is both correct and adheres to a specified format, and a format reward, which encourages the model to present its reasoning chain within designated "think" tags.

Within preference optimization, several works have explored rule-based objectives, though identifying suitable rules is challenging due to the opaqueness of human preferences. Anthropic's Constitutional AI~\citep{bai2022constitutionalaiharmlessnessai} uses a curated set of constitutional principles to guide response revision and preference judgments, but these are not directly used as scalar rewards. DeepMind's Sparrow~\citep{glaese2022improvingalignmentdialogueagents} employs researcher-crafted behavioral rules, collecting rule violation annotations from human raters to train a dedicated rule reward model. By jointly optimizing the policy with both rule-based and preference-based rewards, Sparrow achieves a reduction in rule violations. OpenAI has also investigated rule-based rewards for safety alignment, decomposing policy rules into simple propositions and using them as features in a fitted linear model to construct a reward signal during RL~\citep{NEURIPS2024_c4e380fb}.

Although useful, constructing effective rule sets is costly, requires significant domain expertise, and often demands scenario-specific customization. As a result, rule-based approaches in preference learning remain largely proprietary within industry, with few publicly available rule sets in academic research. 
\begin{figure}
  \centering
  \includegraphics[width=\textwidth]{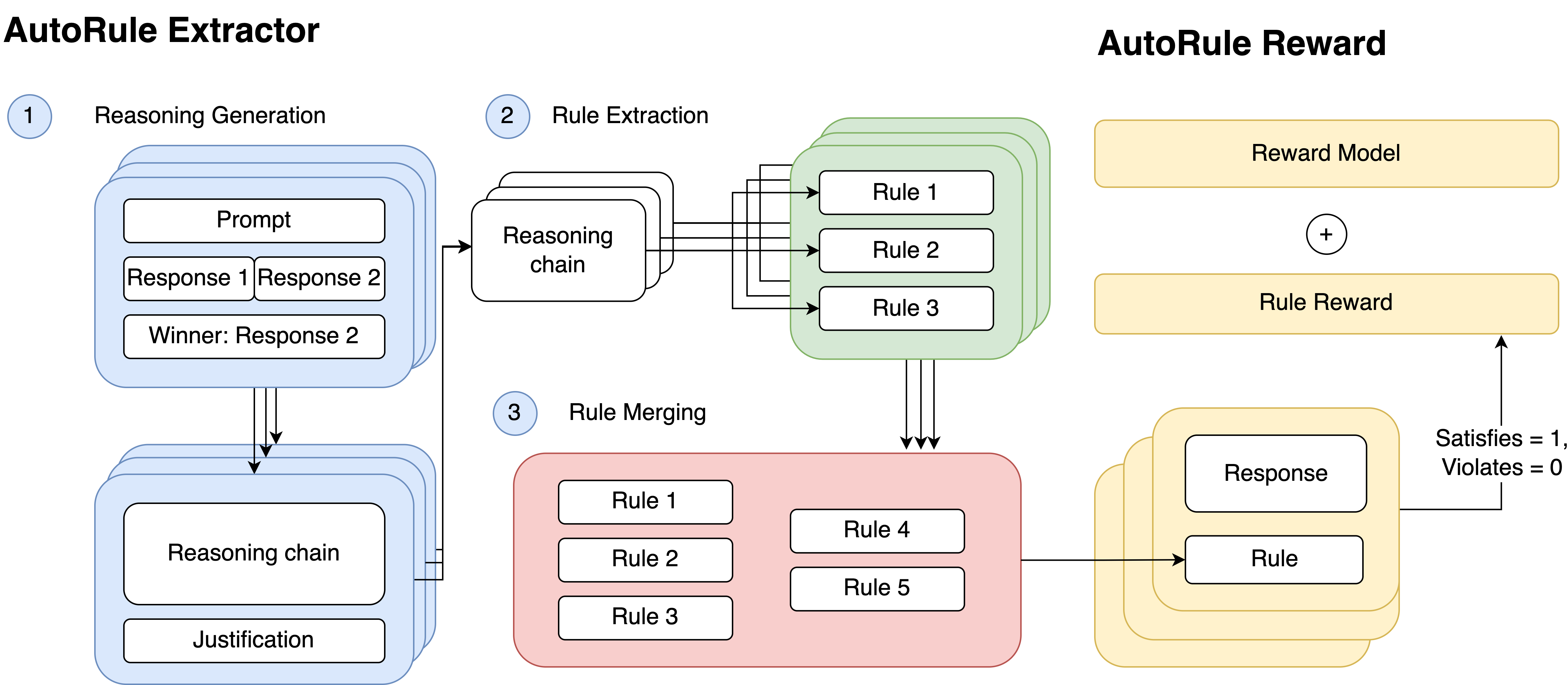}
  \caption{Overview of the \method{} method. }

  \label{fig:overview}
\end{figure}

\section{Methods}
\label{sec:methods}

In this section, we outline the automatic rule extraction process of \method{}, demonstrate how these rules can be used to form a reward score, and how the reward is used in the GRPO formulation. Figure~\ref{fig:overview} provides an overview of the rule extraction and reward computation pipeline.

\subsection{\method{} Extractor}
We denote the language model (LM) as $\pi_\theta$, where a prompt $x$ serves as the state and the next token $t$ as the action, i.e., $t \sim \pi_\theta(\cdot \mid x)$. Unrolling this process over $N$ tokens, the probability of generating an output sequence $y = (y_1, \dots, y_N)$ is given by $\pi_\theta(y \mid x) = \prod_{i=1}^N \pi_\theta(y_i \mid y_{<i}, x)$. For brevity, we write a sampled output as $y \sim \pi_\theta(\cdot \mid x)$.

The automatic rule extraction process in \method{} consists of three main stages, each leveraging a reasoning language model $\pi_\phi$ that decomposes a response $y$ into an output $o$ and an associated reasoning trace $r$, i.e., $(o, r) \sim \pi_\phi(\cdot \mid x)$.


\paragraph{Reasoning Generation.} To guide the reasoning model toward producing a coherent, step-by-step reasoning chain suitable for rule extraction, we prompt it to justify why a selected response is preferable. Given a preference dataset $\mathcal{D}_\text{prefs} = \left\{(x^{(1)}, y_c^{(1)}, y_r^{(1)}), \ldots, (x^{(N)}, y_c^{(N)}, y_r^{(N)}) \right\}$, we present the reasoning model with either $x_1 = \text{prompt}(x, y_c, y_r, 1)$ or $x_2 = \text{prompt}(x, y_r, y_c, 2)$, chosen randomly to vary candidate ordering. Each prompt requests a justification for the model's output. For every example $i$, we extract the reasoning trace $r^{(i)}$ from the model's generation $(o^{(i)}, r^{(i)}) \sim \pi_\phi(\cdot \mid x)$, resulting in a collection of reasoning chains $RC = \left\{r^{(1)}, \ldots, r^{(N)}\right\}$. The prompts used for this step as well as the remaining steps are displayed in Appendix \ref{sec:prompts}. 

\paragraph{Rule Extraction.} Next, we extract explicit rules from each individual reasoning chain. For every reasoning chain $r^{(i)} \in RC$, we prompt the reasoning model with $x = \text{prompt}(r^{(i)})$ to elicit the underlying rules that justify the preference. The model outputs a set of rules $R^{(i)}$ for each $r^{(i)}$, i.e., $R^{(i)}, r \sim \pi_\phi(\cdot \mid x)$. We aggregate these across all examples to obtain the overall rule set:
\[
RS = \bigcup_{i=1}^{N} R^{(i)}
\]
By leveraging the reasoning model in this way, we aim to systematically decompose complex reasoning traces into precise, actionable rules. Extracting rules individually from each reasoning chain also simplifies the model's task which should promote higher-quality and more interpretable rule sets.

\paragraph{Rule Merging.} Given the substantial number of rules extracted from the training set, it is essential to merge rules to ensure computational efficiency during training. To address redundancy and overlap, we prompt the reasoning model to merge rules. To do this, we mention in the prompt that the rules provided might have duplicate or semantically similar rules. We then instruct the model to identify and consolidate rules within $RS$ so that no duplicates or similar rules remain. This results in a refined and compact set of merged rules:
\[
MR, r \sim \pi_\phi(\cdot \mid \text{prompt}(RS))
\]
where $MR$ denotes the final set of merged rules. Empirically, this merging process substantially reduces redundancy, typically compressing the rule set to just $1$--$2\%$ of its original size. This significantly improves the efficiency of the rule-based reward calculation process.

\subsection{\method{} Reward}
\label{sec:method-reward}

To construct rule-based rewards for use in the RL objective, we employ LLM-as-a-judge verifiers, denoted as $V_\theta$. Given a prompt $x$, a response $y$, and each extracted rule $\text{rule}_i \in MR$, the verifier provides a rule score $s_{i} \sim V_\theta(\cdot \mid \text{prompt}(x, y, \text{rule}_i))$. We constrain the rule scores to binary values, $s_{i} \in \{0, 1\}$. The \method{} reward $r_{RA}$ is then defined as the mean rule satisfaction across all $K = |MR|$ rules:
\[
r_{RA}(x, y) = \frac{1}{K} \sum_{i=1}^{K} s_{i}
\]
where each $s_{i}$ is obtained as above. The final reward used for training combines the rule-based reward $r_{RA}$ with the standard reward model score $r_\theta$ and KL penalty (exact KL penalty formulation in Appendix \ref{sec:kl}):
\[
r_{\text{total}}(x, y) = r_{RA}(x, y) + r_\theta(x, y) -  \beta_{KL}KL_{\text{approx}}
\]

Unlike conventional reward models that assign continuous scores reflecting subtle preference distinctions, our verifier $V_\theta$ is tasked solely with determining whether each rule is satisfied, producing a binary outcome. This simplification reduces the complexity of the reward modeling process, making the verifier less susceptible to erroneous judgments, mitigating the risk of reward hacking.

\subsection{\method{} RL Stage}
\label{sec:method-grpo}
\method{} uses the GRPO algorithm \cite{shao2024deepseekmathpushinglimitsmathematical} for the RL stage of the preference alignment, using $r_{\text{total}}$ as the reward signal. GRPO is a policy optimization algorithm that uses the relative rewards from a group of outputs to determine advantage estimates, removing the need for a separate value model used in traditional PPO~\citep{DBLP:journals/corr/SchulmanWDRK17}, increasing memory and computational efficiency. Formally, GRPO utilizes a group of outputs and computes their rewards, consolidating them into a reward vector $\mathbf{r} = \{r_1, \ldots, r_n\}$. Then, it computes advantage estimates for a particular output $i$ as
\[\hat{A}_i = \frac{r_i - \text{mean}(\mathbf{r})}{\text{std}(\mathbf{r})}\]
This advantage estimate is then used in the following clipped surrogate objective \cite{DBLP:journals/corr/SchulmanWDRK17}:
\[L(w) = \mathbb{E}_{(x,y) \sim \mathcal{D}_{w_{old}}} \left[\min\left(\frac{\pi_w(y\mid x)}{\pi_{w_{old}}(y\mid x)}\hat{A}, \text{clip}\left(\frac{\pi_w(y\mid x)}{\pi_{w_{old}}(y\mid x)}, 1 - \epsilon, 1 + \epsilon\right)\hat{A}\right)\right]\]
where $\epsilon$ is a clipping hyperparameter and $\frac{\pi_w(y\mid x)}{\pi_{w_{old}}}$ is the likelihood ratio.

In summary, \method{} introduces an automated, reasoning-chain-based rule extraction framework that can generate precise and actionable alignment rules, thereby eliminating the need for manual rule engineering. Furthermore, by leveraging LLM-as-a-judge verifiers that provide binary rule satisfaction judgments, our approach simplifies reward modeling compared to conventional continuous reward models, which helps mitigate reward hacking and enhances the reliability of preference alignment.
\section{Experimental Methodology}
\label{sec:experiments}

\textbf{Dataset.}
We use the UltraFeedback-Binarized dataset (referred to as UltraFeedback), a binarized version of UltraFeedback \cite{cui2024ultrafeedback}, which contains nearly 64K pairwise preference annotations across diverse model types and instructions.  For training, we select a filtered subset of 33K examples (details in Appendix~\ref{sec:filtering}). We also use the MT-Bench human judgment dataset (referred to as MT-Bench) \cite{NEURIPS2024_c4e380fb}, which provides expert preference annotations on multi-turn questions.

\textbf{Evaluation Metrics.}
We report win rate on the UltraFeedback-Binarized test split, using GPT-4o as an automatic judge with randomized candidate and reference response order. We also evaluate on MT-Bench (using a GPT-4 judge) and AlpacaEval 2.0 \cite{dubois2024lengthcontrolled}. For \method{}, AlpacaEval 2.0 and UltraFeedback win rate are measured on a model trained with rules from UltraFeedback. For MT-Bench, we split the 80 questions into 40 for training \method{} and 40 for testing (5 per category for each split).

\textbf{Rule Extraction.}
We use Deepseek-R1 \cite{deepseekai2025deepseekr1incentivizingreasoningcapability} to generate reasoning chains for automatic rule extraction. For the LLM-as-a-judge verifier, we use Llama-3-8B-Instruct \cite{grattafiori2024llama3herdmodels} for computational efficiency compared to the larger Deepseek-R1 model. To extract rules, we sample 256 random examples from the UltraFeedback training split and for MT-Bench, we use the 40-question training split and sample up to 8 examples per question for training, or all available if fewer.

\textbf{Baselines.}
We compare to several baselines: (1) RLHF using PPO (``RLHF''), (2) GRPO with base reward and no hyperparameter tuning (``GRPO''), (3) GRPO with a length penalty (``GRPO + Length Penalty'', shortened as LP), and (4) GRPO with length-driven hyperparameter tuning (``GRPO + Length Control'', shortened as LC). All baselines use the same learned reward model.

\textbf{\method{} Model.}
For \method{}, we use a scaled rule-based reward $r_{RA}$:
\[
r_{RA'} = \alpha r_{RA} + \beta
\]
with $\alpha = 10$ and $\beta = -7.5$, aligning the rule-based reward magnitude with the learned reward model for stable training. The verifier prompt is modified so $s_i = 1$ only if the response is concise and fully satisfies the extracted rule.

\textbf{Implementation Details.}
All models are initialized from the same SFT and reward model checkpoints for comparability. The SFT checkpoint is obtained by fine-tuning Llama-3-8B on preferred responses from filtered UltraFeedback-Binarized. The reward model is initialized from this SFT checkpoint and further fine-tuned on preference annotations from the filtered UltraFeedback-Binarized training split. Actor, critic, and value networks (where applicable) are initialized from the SFT checkpoint. Training uses OpenRLHF \cite{hu2024openrlhfeasytousescalablehighperformance}, an open-source RLHF framework. Hyperparameters and further details are in Appendix~\ref{sec:training}, and asset URLs are available in Appendix~\ref{sec:licenses}.

\section{Evaluation Results}
\label{sec:results}

In this section, we evaluate \method{} on measures of rule quality, model performance, and reward hacking mitigation. We then analyze a few ablations of \method{} as well as a case study of the rules extracted.

\subsection{Rule quality}
\begin{figure}[t]
  \centering
    \centering
     \begin{subfigure}[t]{0.24\linewidth}
      \includegraphics[width=\linewidth]{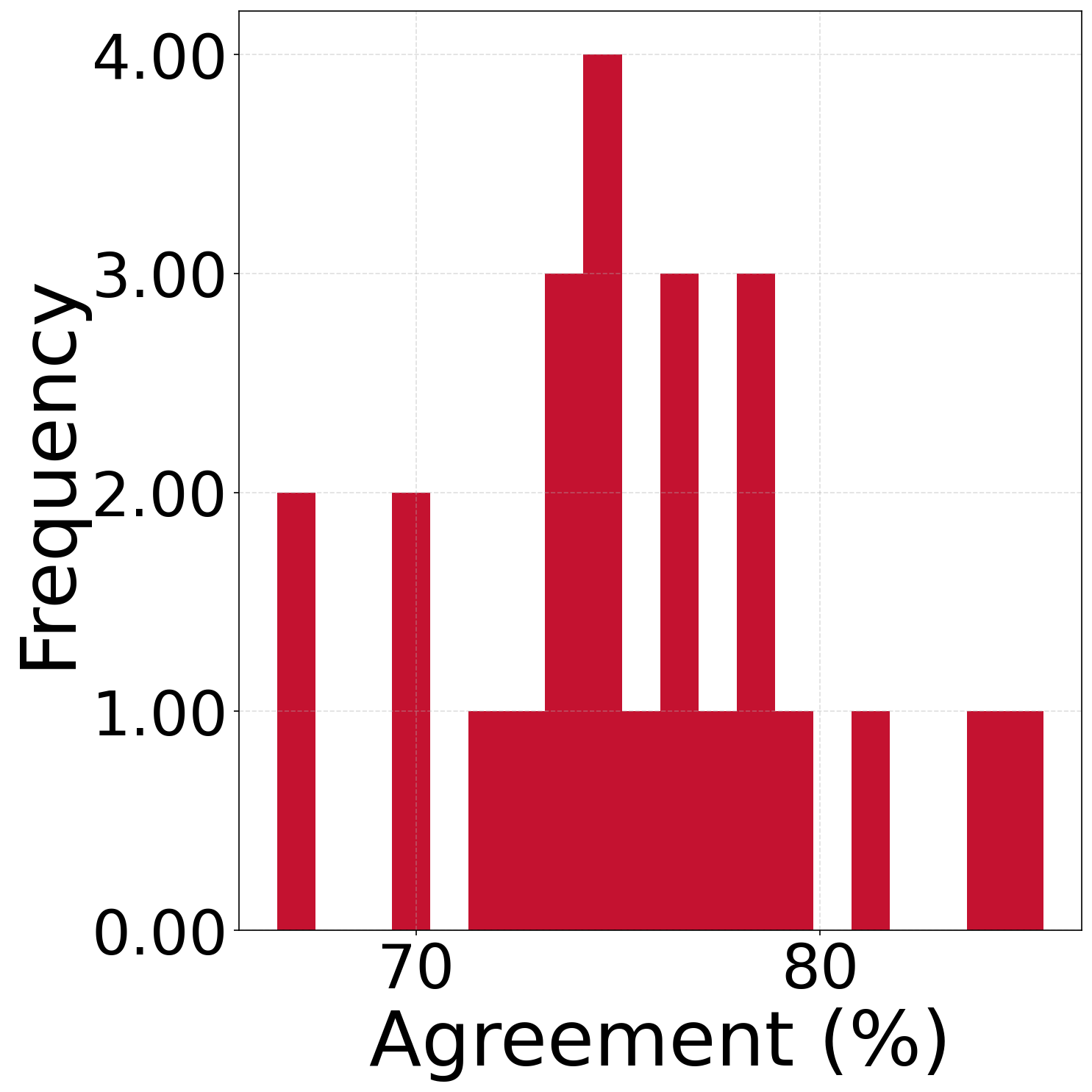}
      \caption{UltraFeedback}
      \label{fig:uf-indv-rule}
    \end{subfigure}
    \begin{subfigure}[t]{0.24\linewidth}
      \includegraphics[width=\linewidth]{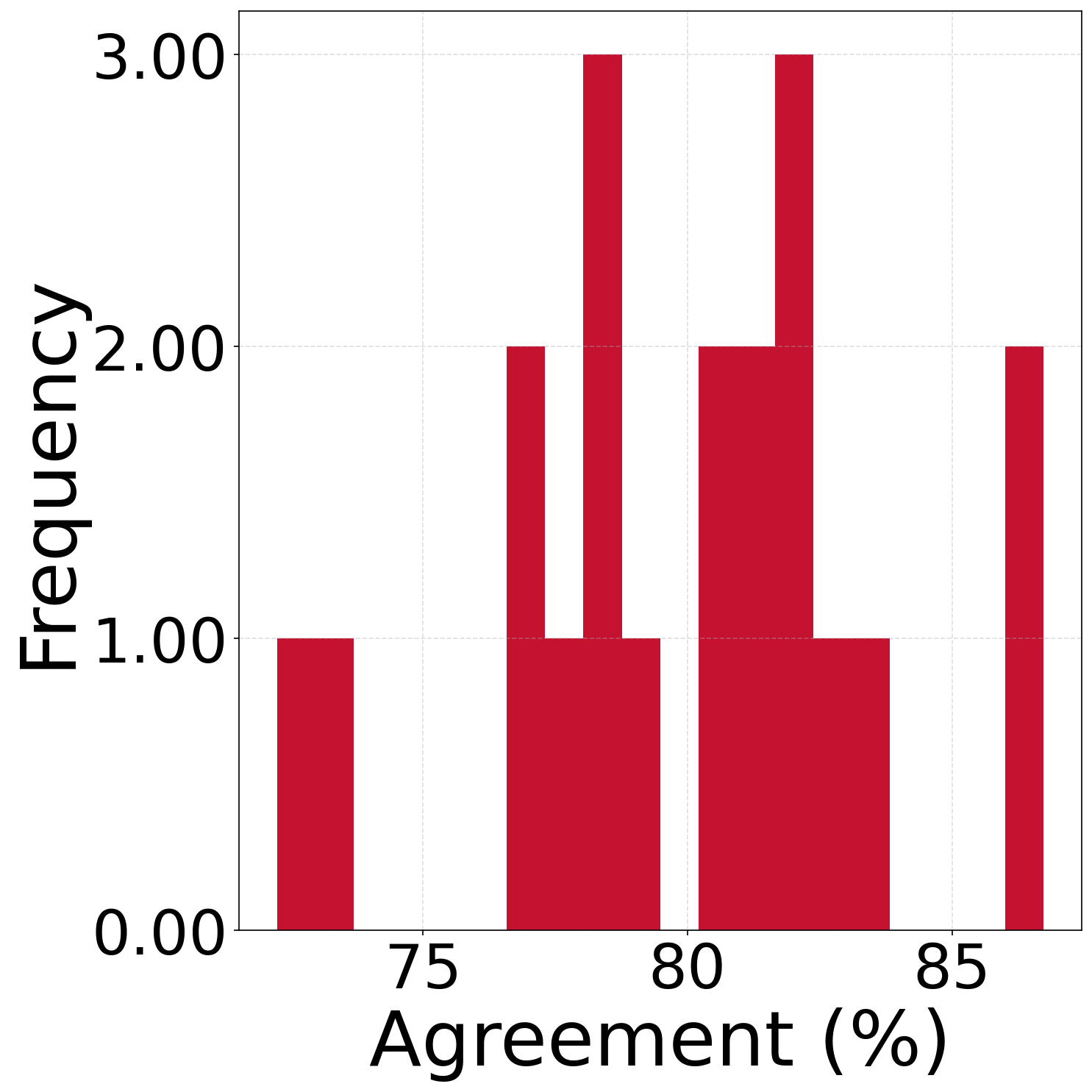}
      \caption{MT-Bench}
      \label{fig:mt-indv-rule}
    \end{subfigure}
    \begin{subfigure}[t]{0.24\linewidth}
      \includegraphics[width=\linewidth]{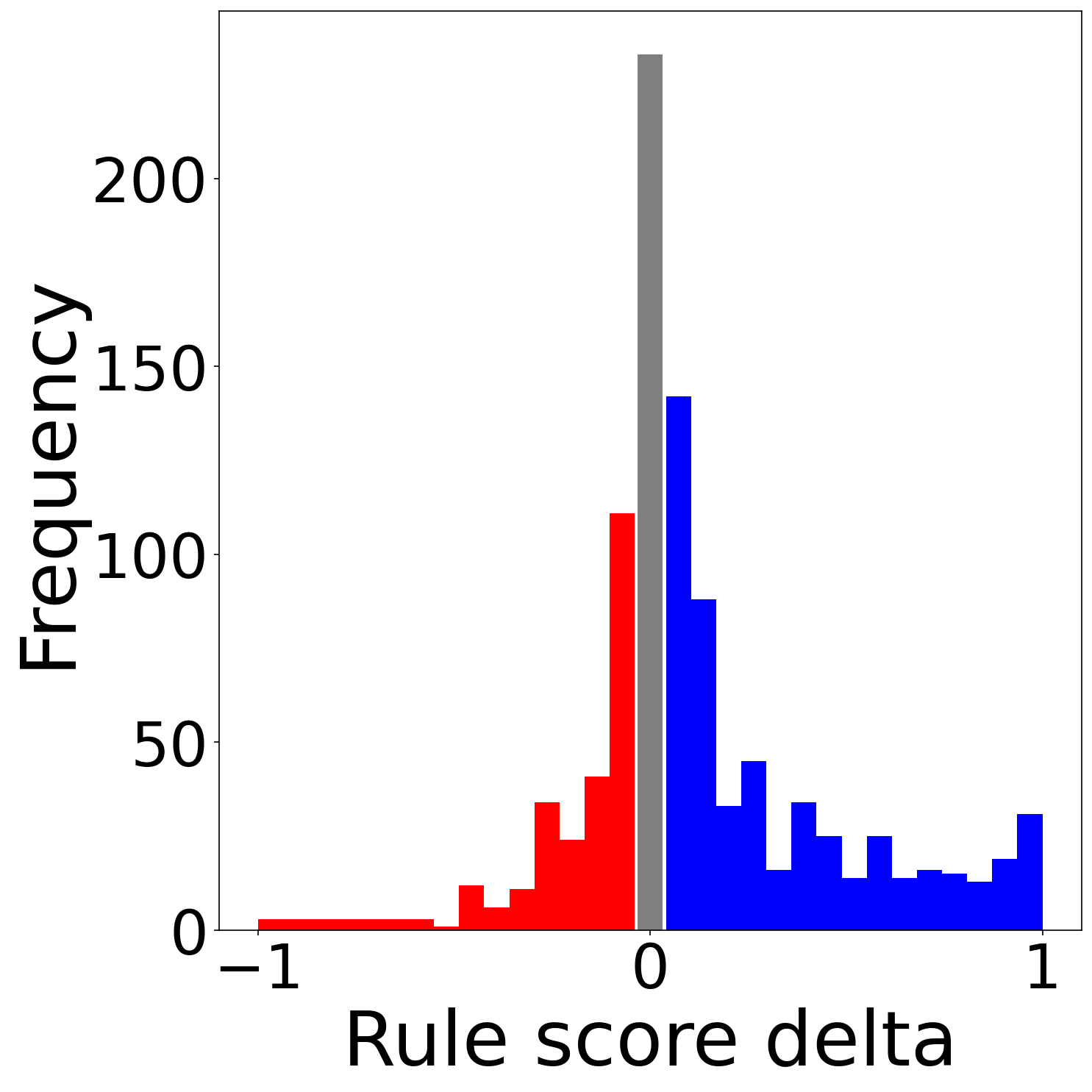}
      \caption{UltraFeedback}
      \label{fig:uf-delta}
    \end{subfigure}
    \begin{subfigure}[t]{0.24\linewidth}
      \includegraphics[width=\linewidth]{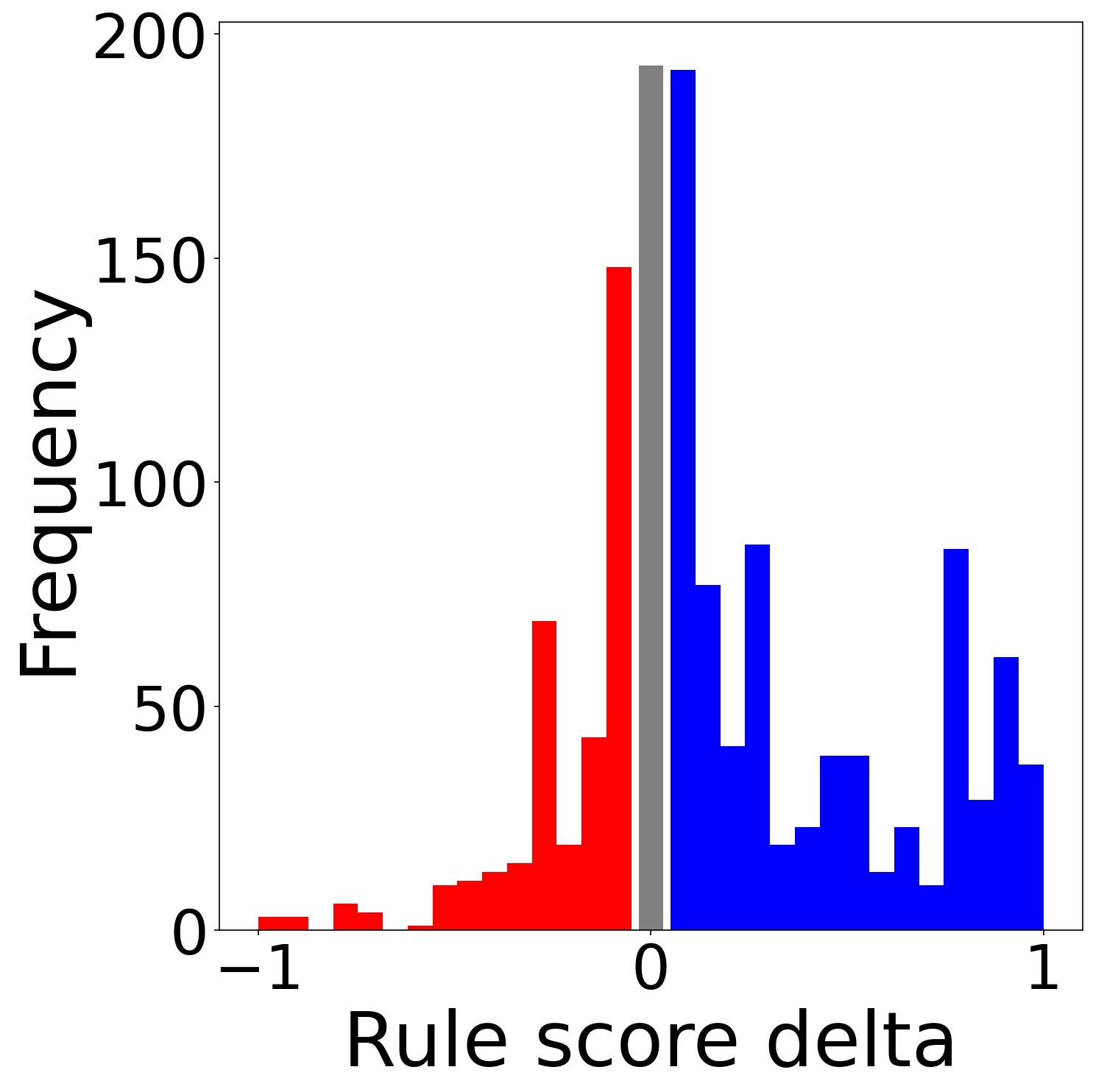}
      \caption{MT-Bench}
      \label{fig:mt-delta}
    \end{subfigure}
    \caption{Distributions of individual rule agreement in Figures \ref{fig:uf-indv-rule} and \ref{fig:mt-indv-rule}, rule score delta distributions for UltraFeedback-extracted and MT-Bench-extracted rules in Figures \ref{fig:uf-delta} and \ref{fig:mt-delta}, respectively. }
    \label{fig:rule-agreements}
\end{figure}

All extracted rules are displayed in Appendix \ref{sec:rules}. We assess rule agreement on 1,024 UltraFeedback test examples and the full MT-Bench human judgment split. For individual rules, agreement is measured as the fraction of response pairs where the rule's preference matches the ground-truth label, considering only pairs where the rule distinguishes between responses. We also report the distribution of aggregate rule reward deltas (difference between chosen reward and rejected reward).

Figures \ref{fig:uf-indv-rule} and \ref{fig:mt-indv-rule} presents the distributions for individual rule agreement. We observe that individual rules from both rule sets are in good agreement with the ground-truth preferences in the datasets. Furthermore, the mean agreement with MT-Bench rules surpasses that with UltraFeedback, indicating that automatic rule extraction seems to be more effective on human-annotated data compared to LLM-generated preferences.  

Figures \ref{fig:uf-delta} and \ref{fig:mt-delta} show the distributions of rule score deltas. The distribution of rule score deltas reveals a heavier tail on the positive side, showing that more chosen responses recieve higher scores. The positive tail becomes heavier as the magnitude increases, implying that larger magnitudes of deltas are associated with more reliable preference alignment.

In addition to rule agreement, we also conducted a small experiment to assess the determinism of the rules by running verifier inference with a temperature of 1.0, 100 times on 20 UltraFeedback test-set responses for the UltraFeedback-extracted rules and 16 MT-Bench test-set responses for the MT-Bench-extracted rules. Using a determinism score calculated as $(\max(\# \text{Yes}, \# \text{No})/(\# \text{Yes} + \# \text{No}))$, where Yes/No indicates the answer to rule satisfaction, we obtained average determinism scores of 83.6\% and 82.5\% for UltraFeedback and MT-Bench-extracted rules, respectively. These results suggest that the rules exhibit a high degree of consistency and that the verifier provides reliable judgments, supporting their suitability for reward formulation.

\subsection{Model Performance}

\begin{table}[t]
\centering
\small
\caption{Main evaluation results on UltraFeedback win-rate, AlpacaEval 2.0, and MT-Bench, all methods are conducted on the \textsc{LlaMA 3 8B} base model.}
\begin{tabular}{lcccc}
\toprule
 & \multicolumn{1}{c}{UF WR} & \multicolumn{2}{c}{AlpacaEval2.0 LC WR (WR)} & \multicolumn{1}{c}{MT-Bench}\\
 \cmidrule(r){2-2} \cmidrule(l){3-4}  \cmidrule(l){5-5} 
\textbf{Methods} & \textbf{vs SFT} (\%) & \textbf{vs SFT} (\%) &  \textbf{vs GPT-4} (\%) & \textbf{Avg (Turn 1/Turn 2)} \\
    \cmidrule(r){1-5}
SFT & -- & -- & 10.8 (7.2) & 6.40 (6.98/5.83) \\
RLHF & 67.6 & 66.3 (67.0) & 15.2 (11.1) & 7.09 (7.41/6.78)\\
\midrule
GRPO & 75.9 & 72.7 (82.2) & 15.1 (16.1) & 7.68 (7.98/7.38)\\
GRPO + Length Control & 75.9 & 66.1 (80.2) & 16.8 (16.8) & 7.40 (7.45/7.35)\\
GRPO + Length Penalty & 76.1 & 71.0 (76.6) & 16.2 (12.5) & 7.29 (7.58/7.00) \\
\midrule
\method{} & \textbf{77.2} & \textbf{77.0} (83.3) & \textbf{21.6} (18.6) & \textbf{7.85} (7.88/7.83)\\
\bottomrule
\end{tabular}
\label{tab:model-eval}
\end{table}

Table \ref{tab:model-eval} presents a comprehensive comparison of baseline and \method{} models across multiple evaluation metrics, including UltraFeedback win rate, AlpacaEval 2.0 length-controlled win rate (LC WR)/regular win rate (WR), and MT-Bench performance. Our results demonstrate that rule-based reward models are effective within their respective training domains. Specifically, when trained with rules extracted from UltraFeedback, \method{} achieves a 1.7\% relative improvement in UltraFeedback win rate over the baselines, indicating that the extracted rules successfully capture important aspects of human preference in this dataset.

For MT-Bench, we evaluate \method{} using rules derived from a curated subset of 40 multi-turn prompts. Notably, the model exhibits a 6.1\% relative gain in Turn 2 performance compared to the baseline, demonstrating the effectiveness of rule-based supervision for complex, multi-turn interactions.

Beyond in-domain performance, our findings indicate that rule-based approaches exhibit superior generalization to out-of-distribution tasks relative to conventional baselines. On AlpacaEval 2.0, \method{} attains a 5.9\% relative improvement in length-controlled win rate against the SFT baseline, and a 28.6\% improvement against GPT-4 Turbo, highlighting the robustness of rule-based rewards in mitigating length bias and promoting substantive response quality. These results collectively suggest that \method{} not only excels within its training distribution but also transfers effectively to diverse evaluation settings, outperforming both standard and length-bias reduction baselines. 

\subsection{Reward Hacking Mitigation}

\begin{figure}[t]
  \centering
    \centering
        \begin{subfigure}[t]{0.24\linewidth} 
      \includegraphics[width=\linewidth]{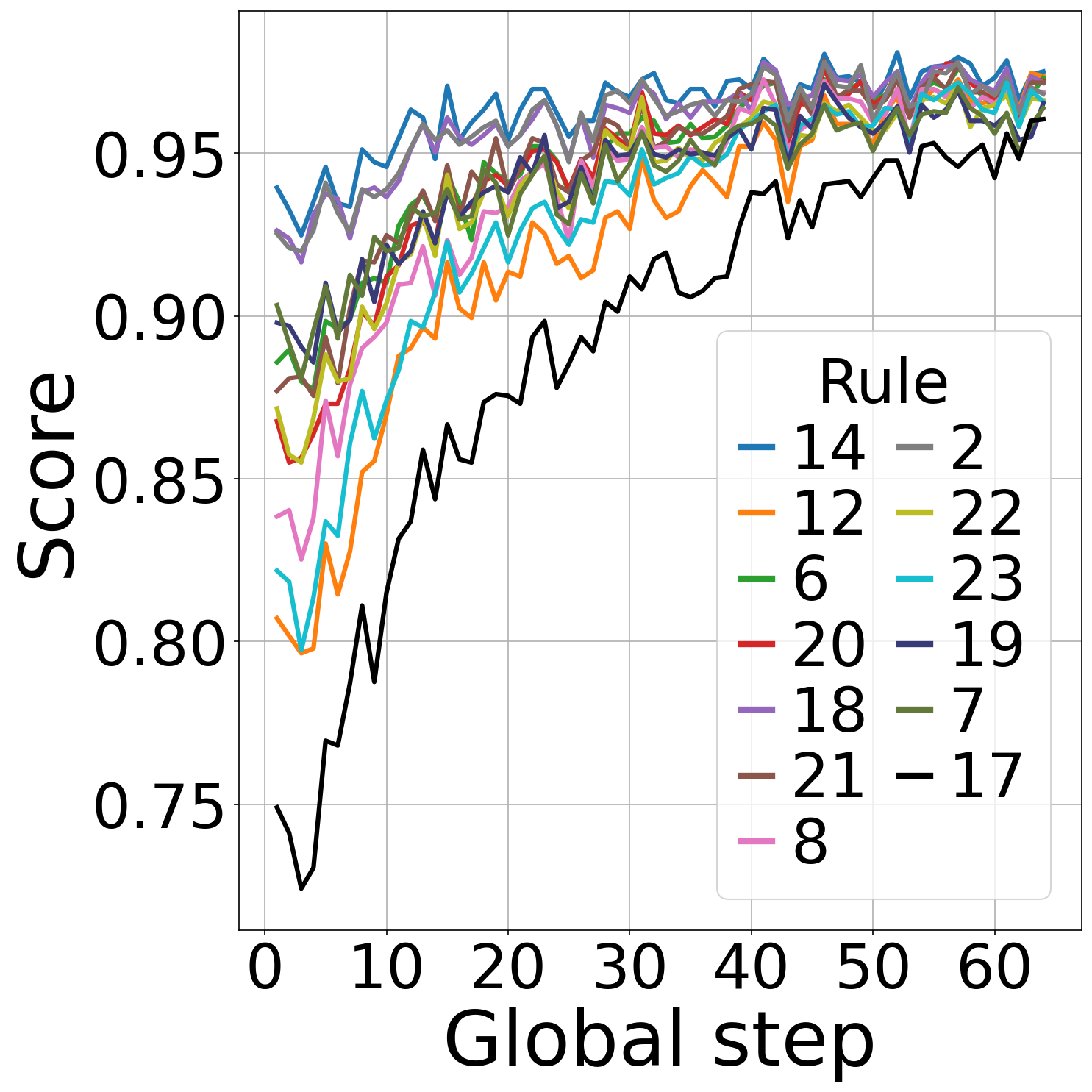}
      \caption{Top half by last score}
      \label{fig:rule-high}
    \end{subfigure}
    \begin{subfigure}[t]{0.24\linewidth} 
      \includegraphics[width=\linewidth]{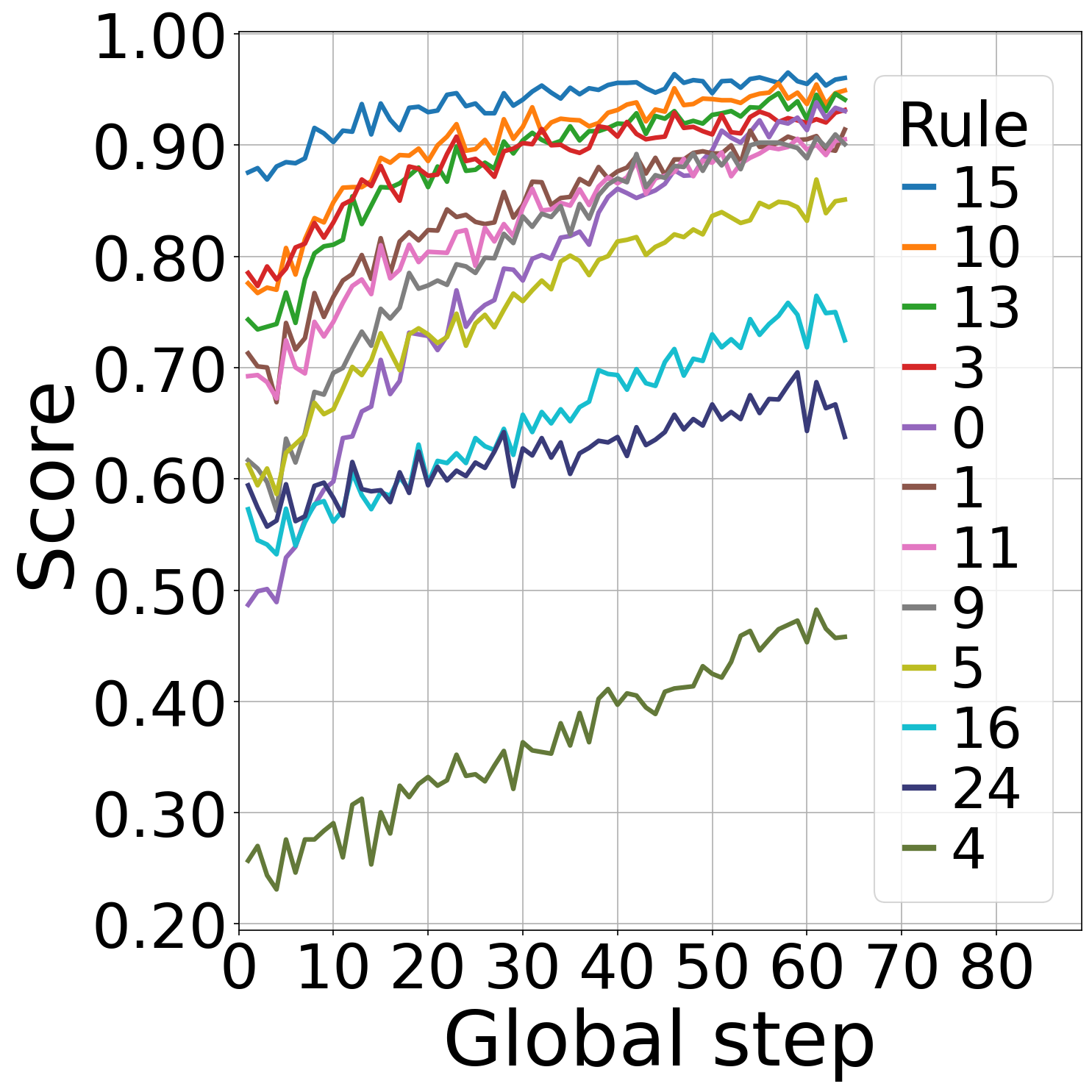}
      \caption{Bot. half by last score}
      \label{fig:rule-low}
    \end{subfigure}
     \begin{subfigure}[t]{0.24\linewidth} 
    \includegraphics[width=\linewidth]{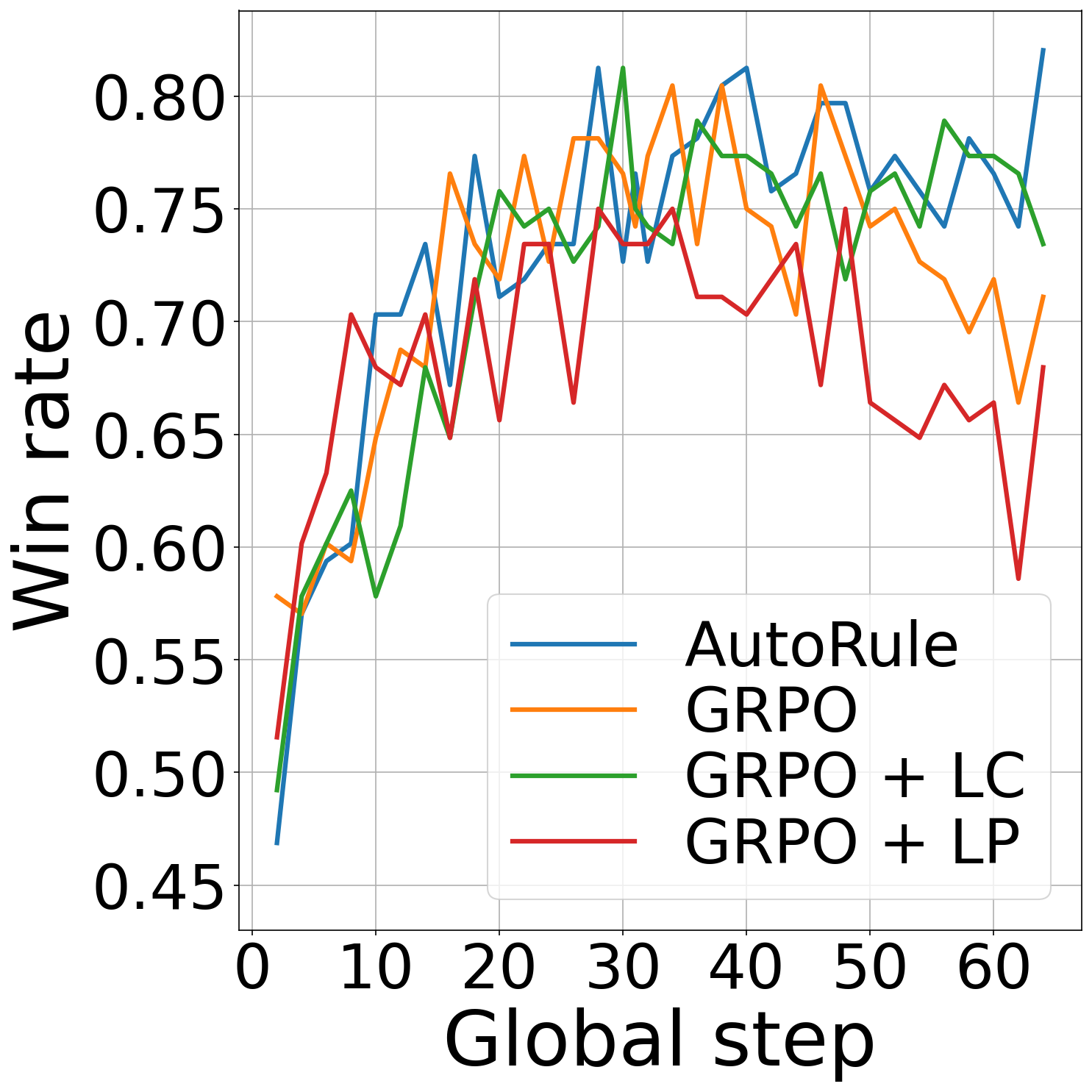}
    \caption{UltraFeedback WR}
    \label{fig:uf-win-rate-vs-step}
    \end{subfigure}
    \begin{subfigure}[t]{0.24\linewidth}
      \includegraphics[width=\linewidth]{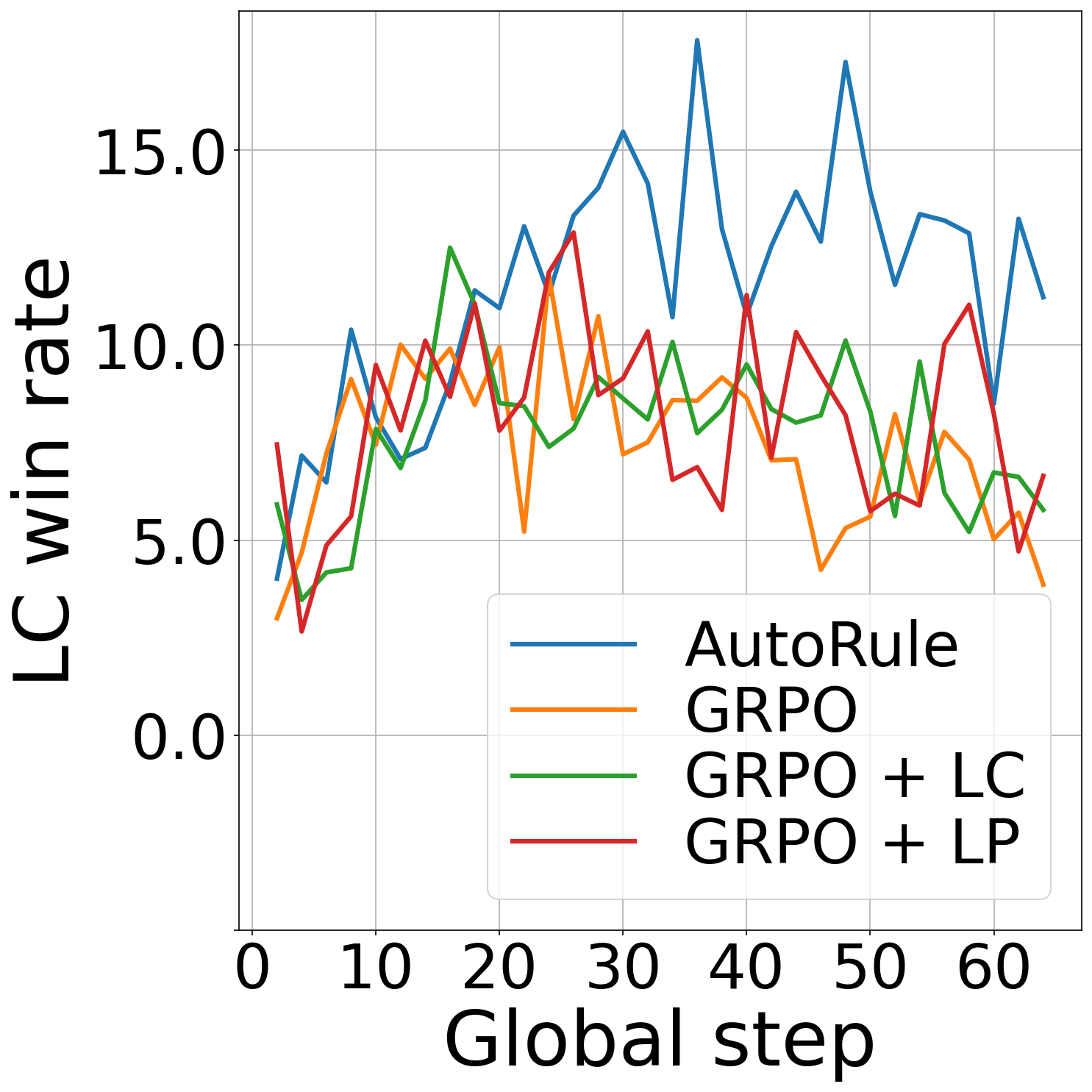}
      \caption{AlpacaEval 2.0 LC WR}
      \label{fig:alpaca-win-rate-vs-step}
    \end{subfigure}
    \caption{Average individual rule scores of \method{} for two episodes, split into two figures (\ref{fig:rule-high} and \ref{fig:rule-low}), evaluation of \method{} and GRPO baselines for two episodes in Figures \ref{fig:uf-win-rate-vs-step} and \ref{fig:alpaca-win-rate-vs-step}.}
    \label{fig:rules-win-rates}
\end{figure}
To systematically evaluate reward hacking, we monitor model performance throughout training to detect any degradation indicative of overfitting to the reward signal. We conduct four experimental runs: three baselines and one instance of \method{} trained with UltraFeedback-derived rules, each for two episodes (i.e., two full passes over the dataset). Models are checkpointed every two steps.

As an initial verification, we report the mean individual rule scores as a function of global step in Figures~\ref{fig:rule-high} and~\ref{fig:rule-low}. The observed upward trajectory across all rules demonstrates that the model is effectively optimizing for the rule-based reward signal. These results validate the use of these training runs for subsequent analysis of reward dynamics under the \method{} framework.

For each checkpoint, we evaluate UltraFeedback win rate relative to the SFT checkpoint and AlpacaEval 2.0 length-controlled win rate against GPT-4, both with only a subset of 128 examples. Figure~\ref{fig:uf-win-rate-vs-step} depicts UltraFeedback win rate as a function of global step. Initially, both baseline and \method{} models achieve similar win rates; however, after step 52, the GRPO and GRPO + LP baselines exhibit declining performance, whereas GRPO + LC and \method{} maintain consistently high win rates. 

For out-of-distribution generalization, Figure~\ref{fig:alpaca-win-rate-vs-step} shows AlpacaEval 2.0 win rate against the global step. Here, \method{} consistently outperforms all GRPO baselines, including GRPO + LC, achieving an improvement of roughly 5 percentage points after two episodes.
While GRPO + LC mitigates reward hacking for in-distribution data, rule-based rewards provide robustness against reward hacking for both in- and out-of-distribution settings. 

\subsection{Ablation Study}
 
\begin{table}[t]
\centering
\small
\caption{Ablation results on UltraFeedback win-rate and AlpacaEval 2.0 using \method, sll methods are conducted on the  Llama-3-8B base model.}

\label{tab:ablation}
\begin{tabular}{lccc}
\toprule
 &\multicolumn{1}{c}{UF WR} & \multicolumn{2}{c}{AlpacaEval2.0 LC WR (WR)}\\
 \cmidrule(r){2-2} \cmidrule(l){3-4} 
\textbf{Model} & \textbf{vs SFT} (\%) & \textbf{vs GPT-4} (\%) & \textbf{vs SFT} (\%) \\
    \cmidrule(r){1-4}
Original (UF-trained) & \textbf{77.2} & \textbf{77.0} (83.3) & \textbf{21.6} (18.6)\\
w/o Scaling, Concise &  75.7 & 68.6 (82.5) & 14.5 (17.8) \\
w/o Concise & 74.6 & 65.2 (82.4) & 16.5 (21.6) \\
\bottomrule
\end{tabular}
\end{table} 

To better understand the contributions of individual components within our framework, we conduct a ablation study focusing on two critical aspects: reward scaling and conciseness constraints. Specifically, we consider the following variants: (1) a model trained without rule-based reward scaling, where the scaling parameters are set to $\alpha = 1, \beta = 0$ (denoted as "w/o Scaling"), and (2) a model in which the verifier prompt is modified to omit references to conciseness, thereby removing explicit guidance for concise responses, along with no scaling (denoted as "w/o Scaling, Concise"). 

The results, summarized in Table \ref{tab:ablation}, reveal that the removal of either reward scaling or conciseness guidance leads to a consistent decline in both UltraFeedback win rate and AlpacaEval 2.0 length-controlled win rate. The absence of reward scaling diminishes the model's ability to effectively leverage rule-based supervision, while omitting conciseness constraints results in responses that are less aligned with human preferences for brevity and clarity. These findings underscore the importance of both scaling the rule reward and explicitly encouraging concise responses within the \method{} framework.

\subsection{Rule Analysis}

\paragraph{Reasoning VS. Justification Rules.} We next investigate the impact of the extraction medium on rule quality and downstream performance. Specifically, we compare rules extracted directly from model justifications (i.e., the immediate output of Deepseek-R1 following the reasoning chain-of-thought) to those extracted from the reasoning chain.

\begin{figure}[t]
  \centering
\small
  \begin{minipage}[t]{0.69\textwidth}
  \vspace{0pt}
\centering
\captionof{table}{Comparison of extraction mediums on UltraFeedback win-rate and AlpacaEval 2.0, methods are conducted on the Llama-3-8B base model.}
\begin{tabular}{lccc}
\toprule
 & \multicolumn{1}{c}{UF WR} 
 & \multicolumn{2}{c}{AlpacaEval2.0 LC WR (WR)}\\
 \cmidrule(l){2-2}  \cmidrule(l){3-4} 
\textbf{Method} & \textbf{vs SFT} (\%) & \textbf{vs SFT} (\%) & \textbf{vs GPT-4} (\%) \\
    \cmidrule(r){1-4}
Best Baseline & 76.1 & 72.7 (82.2) & 16.8 (16.8) \\
\midrule
Reasoning Chain &\textbf{77.2} & \textbf{77.0} (83.3) & \textbf{21.6} (18.6)\\
Justification  & 75.9 & 75.9 (82.2) & 19.7 (16.5) \\
\bottomrule
\end{tabular}
\label{tab:exp-vs-reason}
  \end{minipage}
  \begin{minipage}[t]{0.30\textwidth}
  \vspace{0pt}
    \centering
     \begin{subfigure}[t]{0.78\linewidth}  
      \includegraphics[width=\linewidth]{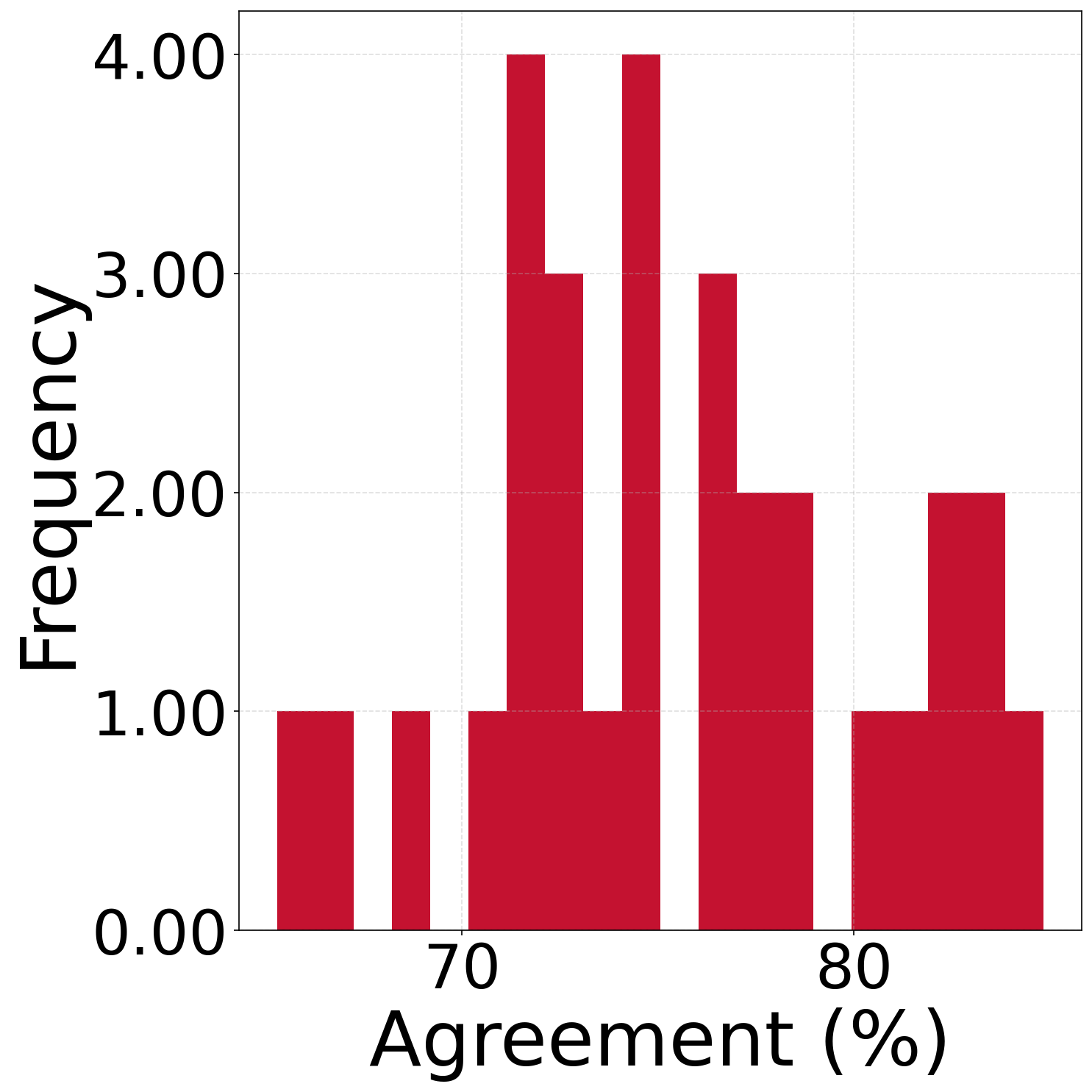}
    \end{subfigure}
      \captionof{figure}{Justification per-rule agreement distribution.}
      \label{fig:exp-indv-rules}
  \end{minipage}
\end{figure}

Table \ref{tab:exp-vs-reason} reports the model performance results of comparing these two extraction mediums. Additionally, the individual rule agreement distribution for model justification is displayed in Figure \ref{fig:exp-indv-rules}. While the rule agreement distribution is similar to reasoning chain, we find that extracting rules from the reasoning chain still yields substantially higher UltraFeedback win rates and AlpacaEval 2.0 length-controlled win rates. This suggests that reasoning chain contains information that allows for the extraction of better rules. We suspect that this is because the reasoning chain provides more specific and actionable guidance for rule formulation, whereas justifications tend to be less detailed and more generic, leading to diminished downstream performance, and justify this via a case study in Appendix \ref{sec:case-study}.

\begin{table}[t]
  \centering
  \small
  \caption{Top 6 unique and similar rules from the UltraFeedback and MT-Bench rule sets, ranked by maximum agreement (\%) as the similarity metric; ``unique'' rules exhibit low agreement with all rules from the other set, while ``similar'' rules show high agreement with at least one rule from the other set.}
  \label{tab:unique-similar-rules}
  \begin{tabular}{p{43mm}ccp{43mm}cc}
\toprule
\multicolumn{3}{c}{Top unique rules}  & \multicolumn{3}{c}{Top similar rules}\\
\cmidrule(l){1-3} \cmidrule(l){4-6}
\textbf{Rule} & \textbf{Data} & \textbf{Max} & \textbf{Rule} & \textbf{Data} & \textbf{Max} \\
\midrule
Ensure conciseness by eliminating redundancy and focusing on core query relevance. & UF & 75.6 & The assistant must maintain logical flow between concepts and preserve essential content in creative adaptations. & MT & 96.1  \\
Balance thoroughness with brevity to prevent information overload while ensuring clarity. & UF & 78.6 & Maintain narrative coherence with source material when discussing plots or characters. & UF & 96.1 \\
The assistant's self-evaluations must critically assess response quality and identify specific improvement areas. & MT & 79.1 & Validate answers against provided context and avoid unsupported extrapolation. & UF & 95.4 \\
Avoid assumptions when ambiguity exists; seek clarification for insufficient context. & UF & 80.6 & The assistant's responses must maintain a professional and approachable tone, adapting to the nature of the user's query. & MT & 95.2  \\
The assistant's responses must provide detailed step-by-step explanations and calculations to ensure correctness and clarity. & MT & 82.2 & Address ethical considerations, legal compliance, and recommend professional consultation when relevant. & UF & 95.2 \\
The assistant's code examples must be complete, functional, and demonstrate separation of concerns (HTML/CSS/JS). & MT & 82.3 & The assistant must avoid contradictions between answers and follow-up explanations while maintaining roleplay consistency. & MT & 94.9 \\
\bottomrule
\end{tabular}
\end{table}

\paragraph{Rule Agreements.} To further investigate the effectiveness of rule extraction, we conduct a comparative analysis of rule sets derived from UltraFeedback and MT-Bench. Specifically, we construct a rule agreement matrix by evaluating all pairs of rules on a test set of 1,024 UltraFeedback examples and the full MT-Bench human judgment test split. Based on this matrix, we identify similar and unique rules according to their agreement scores.

Table \ref{tab:unique-similar-rules} present the top six unique and top six similar rules, respectively, as determined by the maximum rule agreement with rules from the opposing set across both UltraFeedback and MT-Bench examples. Unique rules from UltraFeedback seem to predominantly emphasize conciseness and clarity, while unique rules from MT-Bench are oriented toward handling complex tasks, such as self-evaluation, performing calculations, or providing code examples. This distinction likely reflects the broader topical diversity of UltraFeedback and the specialized, challenging nature of MT-Bench prompts. In contrast, the similar rules shared between the two sets consistently address core aspects of high-quality assistant responses, including logical flow, professional tone, contextual coherence, and answer consistency. Comprehensive rule agreement matrices are provided in Appendix \ref{sec:rule-agreement-matrices} for further reference.
\section{Conclusion}
\label{sec:conclusion}

In this paper, we introduce \method{}, an reasoning chain-based automatic rule extraction mechanism for the utilization of rule-based rewards in language model alignment. We show that the rules extracted from \method{} are in good agreement with the preference dataset, and provide improvements in performance in model evaluations on instruction-following benchmarks. We also demonstrate that the rule-based rewards approach mitigates some levels of reward model overoptimization. Using \method{}, we hope to assist language model researchers in construct and utilize better rule-based rewards in their post-training pipelines.

\bibliography{custom}

\appendix
\newpage
\section{Discussion}

\subsection{Limitations}
\label{sec:limitations}
While our approach shows promising generalization from UltraFeedback to AlpacaEval 2.0, further work will be done to evaluate \method{}'s ability to transfer across a broader range of tasks and domains. Additionally, developing a formal theoretical framework to better understand and improve the mechanisms by which rule-based methods like \method{} mitigates reward hacking remains an important direction for future research.

\subsection{Broader impacts}
\label{sec:broader-impacts}
This work has the potential to advance the development of conversational agents that are both more helpful and less prone to harmful behaviors, by mitigating overoptimization and idiosyncrasies commonly observed in reward model-based approaches. The interpretability afforded by the rule-based framework enhances transparency, enabling researchers and practitioners to better understand and scrutinize the alignment mechanisms governing large language models.

\section{Additional experiment details}

\label{sec:training}

\subsection{Training settings}

Settings used for the SFT, reward model, and RL training are available in Tables \ref{tab:sft-hparams}, \ref{tab:rm-hparams}, and \ref{tab:rlhf-hparams} respectively.

\subsection{Inference parameters}

Inference parameters are displayed in Table \ref{tab:inference}.

\subsection{KL approximation}
\label{sec:kl}
We utilize two versions of KL approximation as implemented in OpenRLHF \cite{hu2024openrlhfeasytousescalablehighperformance}. The first is used for PPO, and the second is used for GRPO.
\begin{align}
&\log\left(\frac{\pi_\phi(y\mid x)}{\pi^{SFT}(y\mid x)}\right)\\
&e^{-\log\left(\frac{\pi_\phi(y\mid x)}{\pi^{SFT}(y\mid x)}\right)} - 1 + \log\left(\frac{\pi_\phi(y\mid x)}{\pi^{SFT}(y\mid x)}\right)
\end{align}

\subsection{Length penalty}
To implement the length penalty, we subtract the following from the reward:
\[\frac{1}{2} \left(\frac{\text{response\_length}}{L}\right) - \frac{1}{2} \]
where $L = 300$ is the target length.

\subsection{GRPO Advantage estimation}

To improve numerical stability, as implemented in OpenRLHF, we utilize a modified version of the advantage estimation formula displayed in Section \ref{sec:method-grpo} as follows:
\[\hat{A}_i = \frac{r_i - \text{mean}(\mathbf{r})}{\text{std}(\mathbf{r}) + 10^{-9}}\]

\subsection{Dataset Filtering}

Following the filtering process and using the code by \cite{fu2025rewardshapingmitigatereward}, to select data for training, we filter and only include the examples where the chosen and rejected responses are both less than 512 tokens, the chosen score is higher than the rejected score, and the word ``confidence" is not in the either response.

\label{sec:filtering}

\begin{table}[H]
\centering
\small
\caption{Supervised fine-tuning  settings.}
\label{tab:sft-hparams}
\begin{tabular}{lp{4.6cm}}
\toprule
\textbf{Setting} & \textbf{Value} \\
\midrule
Base model                    & Llama-3-8B \\
Dataset / split               & UltraFeedback SFT training split \\
Epochs                        & 2 \\
Train / micro batch           & 256 / 2 \\
Learning rate                 & $5\times10^{-6}$ \\
LR scheduler       & Constant with warmup  \\
LR warmup ratio    & 0.1 \\
Adam $\beta_1, \beta_2$             & 0.9, 0.95 \\
Precision                     & bfloat16 \\
Grad-norm clip                & 10 \\
Seed                          & 42 \\
\bottomrule
\end{tabular}
\end{table}

\begin{table}[H]
\centering
\caption{Reward model training settings.}
\label{tab:rm-hparams}
\begin{tabular}{lp{4.6cm}}
\toprule
\textbf{Setting} & \textbf{Value} \\
\midrule
Base checkpoint               & Llama-3-8B SFT checkpoint \\
Loss                          & Pairwise sigmoid \\
Epochs                        & 1 \\
Train / micro batch           & 256 / 2 \\
Learning rate                 & $5\times10^{-6}$ \\
LR scheduler       & Constant with warmup  \\
LR warmup ratio    & 0.1 \\
Adam $\beta_1, \beta_2$             & 0.9, 0.95 \\
Attention               & Flash \\
Precision                     & bfloat16 \\
Grad-norm clip                & 5 \\
Seed                          & 42 \\
\bottomrule
\end{tabular}
\end{table}

\begin{table}[H]
\centering
\caption{RL training settings. \method{} and GRPO + LP uses the same settings as GRPO, except the verifier for \method{} uses a temperature of 0.0.}
\label{tab:rlhf-hparams}
\begin{tabular}{lccc}
\toprule
\textbf{Setting} & \textbf{PPO} & \textbf{GRPO} & \textbf{GRPO + LC}\\
\midrule
Actor init (policy)           & SFT ckpt & SFT ckpt & SFT ckpt \\
Reward / critic init          & RM ckpt  & RM ckpt & RM ckpt \\
Dataset / split               & UF Prefs training split & same & same \\
KL estimator version                  & (1) & (2) & (2) \\
Initial $\beta_{\text{KL}}$   & 0.005 & 0.001 & 0.005 \\
$\lambda$    & 0.95 & 1.00 & 1.00 \\
$\gamma$     & 1 & 1 & 1 \\
Samples per prompt      & 1 & 2 & 2 \\
PTX coefficient               & 0.05 & 0.05 & 0.05 \\
Actor learning rate           & $3\times10^{-7}$ & $5\times10^{-7}$ & $3\times10^{-7}$ \\
Critic learning rate          & $5\times10^{-6}$ & $9\times10^{-6}$ & $9\times10^{-6}$ \\
LR scheduler       & Constant w/ warmup & same & same  \\
LR warmup ratio    & 0.1 & 0.1 & 0.1 \\
Adam $\beta_1, \beta_2$             & 0.9, 0.95 & 0.9, 0.95 & 0.9, 0.95 \\
Rollout batch (total/micro)    & 1024 / 32 & 1024 / 32 & 1024 / 32 \\
Train batch (total/micro)      & 128 / 16 & 128 / 16 & 128 / 16 \\
Temperature / top-$p$             & 0.9 / 0.9 & 1.0 / 1.0 & 0.9 / 1.0 \\
vLLM max new tokens & 1024 & 1024 & 1024 \\
Epochs                         & 1 & 1 & 1 \\
Precision              & bfloat16 & bfloat16 & bfloat16 \\
Attention               & Flash & Flash & Flash \\
Seed                          & 42 & 42 & 42 \\
Grad-norm clip                & 5 & 1 & 1 \\
\bottomrule
\end{tabular}
\end{table}

\begin{table}[H]
\centering
\small
\caption{Inference parameters used, ``preset'' means it is defined by the benchmark.}
\label{tab:inference}
\begin{tabular}{lccc}
\toprule
\textbf{Model (Purpose)} & \textbf{Temperature} & \textbf{Top P} & \textbf{Max new tokens} \\
\midrule
Verifier (Determinism experiment) & 1.0 & 1.0 & 256\\
Verifier (Rule agreement experiment) & 0.0 & 1.0 & 256\\
Trained model (UF win-rate) & 0.9 & 1.0 & 1024\\
Trained model (AlpacaEval 2.0) & 0.9 & 1.0 & 1024\\
Trained model (MT-Bench) & Preset & 1.0 & 1024\\
Deepseek-R1 (Full rule extraction process) & 0.6 & 1.0 & 32768 \\
\bottomrule
\end{tabular}
\end{table}

\newpage
\section{Rules}
\label{sec:rules}

Rules extracted from UltraFeedback (reasoning chain), MT-Bench (reasoning chain), and UltraFeedback (justification) are displayed in Tables \ref{tab:ultra-rules}, \ref{tab:mt-rules}, \ref{tab:exp-rules} respectively.

\begin{table}[H]
\centering
\small
\caption{UltraFeedback rules extracted via the \method{} extraction process. }
\label{tab:ultra-rules}
\begin{tabular}{cp{10cm}c}
\toprule
\textbf{Index} & \textbf{Rule} & \textbf{Agree (\%)} \\
\midrule
0 & The assistant's responses should present explanations in a coherent, step-by-step structure with logical flow, numbered points, and clear sections. & 75.1 \\

1 & When addressing user misconceptions, the assistant must clarify misunderstandings before offering solutions. & 75.9 \\

2 & Translations must use accurate terminology, preserve original tone and structure, and avoid introducing unrelated content. & 79.4 \\

3 & Responses must prioritize technical accuracy, correct formulas, error-free code examples, and validated context alignment. & 76.2 \\

4 & Incorporate vivid sensory details, figurative language, and relatable examples when explicitly requested. & 74.1 \\

5 & Provide actionable advice, practical steps, and concrete implementation strategies tailored to the user's context. & 74.8 \\

6 & Indicate confidence levels while acknowledging uncertainty and limitations when appropriate. & 74.8 \\

7 & Maintain a conversational, empathetic, and professional tone while avoiding overly formal or dismissive language. & 71.4 \\

8 & Integrate cultural sensitivity, domain-specific terminology, and contextual relevance into explanations. & 73.9 \\

9 & Include properly formatted citations, references, and academic conventions when required. & 73.1 \\

10 & Address all components of the user's query comprehensively without omission or tangential content. & 73.6 \\

11 & Avoid assumptions when ambiguity exists; seek clarification for insufficient context. & 69.9 \\

12 & Use illustrative examples of both correct/incorrect approaches to demonstrate concepts. & 78.2 \\

13 & Strictly adhere to user-specified formats, structures, and output requirements. & 70.2 \\

14 & Address ethical considerations, legal compliance, and recommend professional consultation when relevant. & 80.9 \\

15 & Prioritize security measures, error handling, and technical robustness in solutions. & 78.1 \\

16 & Ensure conciseness by eliminating redundancy and focusing on core query relevance. & 67.4 \\

17 & Explain underlying mechanisms, reasoning processes, and cause-effect relationships explicitly. & 74.8 \\

18 & Validate answers against provided context and avoid unsupported extrapolation. & 85.5 \\

19 & Maintain narrative coherence with source material when discussing plots or characters. & 84.3 \\

20 & Structure comparisons, analyses, and recommendations using clear categorization. & 76.1 \\

21 & Anticipate user needs by providing comprehensive details without requiring follow-ups. & 78.6 \\

22 & Preserve specific terms, measurements, and formatting conventions during localization. & 76.3 \\

23 & Use collaborative language and hierarchical organization for complex information. & 77.4 \\

24 & Balance thoroughness with brevity to prevent information overload while ensuring clarity. & 66.6 \\
\bottomrule
\end{tabular}
\end{table}

\begin{table}[H]
\centering
\small
\caption{MT-Bench rules extracted via the \method{} extraction process.}
\label{tab:mt-rules}
\begin{tabular}{cp{10cm}c}
\toprule
\textbf{Index} & \textbf{Rule} & \textbf{Agree} (\%) \\
\midrule
0 & The assistant's responses must provide detailed step-by-step explanations and calculations to ensure correctness and clarity. & 86.7 \\

1 & The assistant's code should avoid unnecessary complexity, handle edge cases, include error handling, and use appropriate data structures. & 80.4 \\

2 & The assistant's responses must maintain a professional and approachable tone, adapting to the nature of the user's query. & 83.8 \\

3 & The assistant's responses must strictly adhere to user-specified formats (e.g., JSON/YAML) with correct syntax and structure. & 72.2 \\

4 & The assistant's explanations should prioritize logical coherence, clarity, and avoidance of redundant or ambiguous content. & 78.2 \\

5 & The assistant must adhere to ethical guidelines by avoiding medical diagnoses and prioritizing user safety in critical situations. & 77.0 \\

6 & Creative outputs must maintain structural integrity (e.g., rhyme schemes, metaphors) while retaining key informational elements. & 78.7 \\

7 & The assistant should proactively address user misunderstandings, anticipate follow-up questions, and provide actionable feedback. & 86.7 \\

8 & The assistant must apply appropriate theoretical principles (e.g., Bayes' theorem) and clarify their relevance to the problem. & 81.7 \\

9 & The assistant's responses should validate assumptions, acknowledge limitations, and use verified data in calculations. & 77.2 \\

10 & The assistant must tailor recommendations to user constraints (e.g., allergies, pregnancy) and cultural context. & 81.6 \\

11 & The assistant's structured outputs should prioritize readability through proper formatting and organizational patterns. & 80.9 \\

12 & The assistant must avoid contradictions between answers and follow-up explanations while maintaining roleplay consistency. & 78.4 \\

13 & The assistant should provide culturally adapted translations of idioms/phrases rather than literal interpretations. & 78.0 \\

14 & The assistant must verify numerical accuracy through step-by-step validation and real-world feasibility checks. & 81.9 \\

15 & The assistant's code examples must be complete, functional, and demonstrate separation of concerns (HTML/CSS/JS). & 79.2 \\

16 & The assistant should address all query components methodically, even if intermediate steps contain errors. & 81.1 \\

17 & The assistant must maintain logical flow between concepts and preserve essential content in creative adaptations. & 82.8 \\

18 & The assistant should prioritize factual accuracy over hypothetical interpretations unless explicitly requested. & 82.1 \\

19 & The assistant's self-evaluations must critically assess response quality and identify specific improvement areas. & 73.6 \\

\bottomrule
\end{tabular}
\end{table}

\begin{table}[H]
\centering
\small
\centering
\small
\caption{UltraFeedback rules extracted on justifications instead of reasoning CoTs.}
\label{tab:exp-rules}
\begin{tabular}{cp{10cm}c}  
\toprule
\textbf{Index} & \textbf{Rule} & \textbf{Agree (\%)} \\
\midrule
0 & The assistant's responses should include concrete examples, actionable insights, and specific applications to explain mechanisms and variables. & 73.1 \\
1 & The assistant's code must handle edge cases, ensure functionality, avoid unsafe practices, and include error handling. & 81.2 \\
2 & Structure explanations logically with step-by-step formats, clear sections, and thematic grouping while maintaining flow. & 72.8 \\
3 & Correct user misconceptions with accurate information using empathetic and polite language. & 80.5 \\
4 & Be concise, avoid redundancy, and prioritize clarity by eliminating unnecessary details. & 65.3 \\
5 & Provide complete, functional code examples with necessary parameters and modular structures. & 77.3 \\
6 & Maintain a neutral, professional tone appropriate to context without unsolicited commentary. & 74.7 \\
7 & Strictly adhere to user instructions without deviation or unwarranted assumptions. & 71.8 \\
8 & Use structured formatting like bullet points and headings for readability and scannability. & 74.8 \\
9 & Address all query components comprehensively with direct answers and relevant context. & 76.7 \\
10 & Validate code functionality, address pitfalls, and ensure integration with existing setups. & 78.3 \\
11 & Anticipate implicit needs while avoiding speculative language beyond provided evidence. & 82.1 \\
12 & Include practical details, alternatives, and implementation steps for real-world application. & 76.2 \\
13 & Ensure technical accuracy, correct terminology, and compliance with domain standards. & 82.8 \\
14 & Avoid tangential topics and focus strictly on core requests without scope creep. & 67.1 \\
15 & Transparently admit limitations and provide actionable alternatives when uncertain. & 70.6 \\
16 & Prioritize ethical responsibility, legal compliance, and cultural sensitivity. & 83.8 \\
17 & Use precise language, avoid jargon, and explain technical terms contextually. & 77.9 \\
18 & Incorporate error handling, reliability checks, and security best practices. & 78.8 \\
19 & Balance brevity with necessary detail, adapting to user's proficiency level. & 68.7 \\
20 & Provide self-contained, compilable code with headers and standard libraries. & 71.4 \\
21 & Maintain logical coherence, avoid contradictions, and ensure factual consistency. & 83.7 \\
22 & Structure narratives chronologically/thematically with clear cause-effect relationships. & 71.5 \\
23 & Use empathetic tone, constructive feedback, and collaborative language. & 74.9 \\
24 & Include quantitative data, contextual reasoning, and measurable outcomes. & 71.9 \\
25 & Offer platform-agnostic solutions unless specific tools are requested. & 73.0 \\
26 & Highlight key takeaways with memorable framing and searchable keywords. & 73.1 \\
27 & Ensure translations preserve meaning, context, and grammatical correctness. & 84.8 \\
28 & Link concepts to real-world impacts, case studies, and stakeholder outcomes. & 74.5 \\
29 & Adopt solution-oriented tone with proactive guidance and troubleshooting tips. & 76.9 \\
\bottomrule
\end{tabular}
\end{table}

\newpage
\section{Rule agreement matrices}
\label{sec:rule-agreement-matrices}

\begin{figure}[H]
  \centering
  \includegraphics[width=\textwidth]{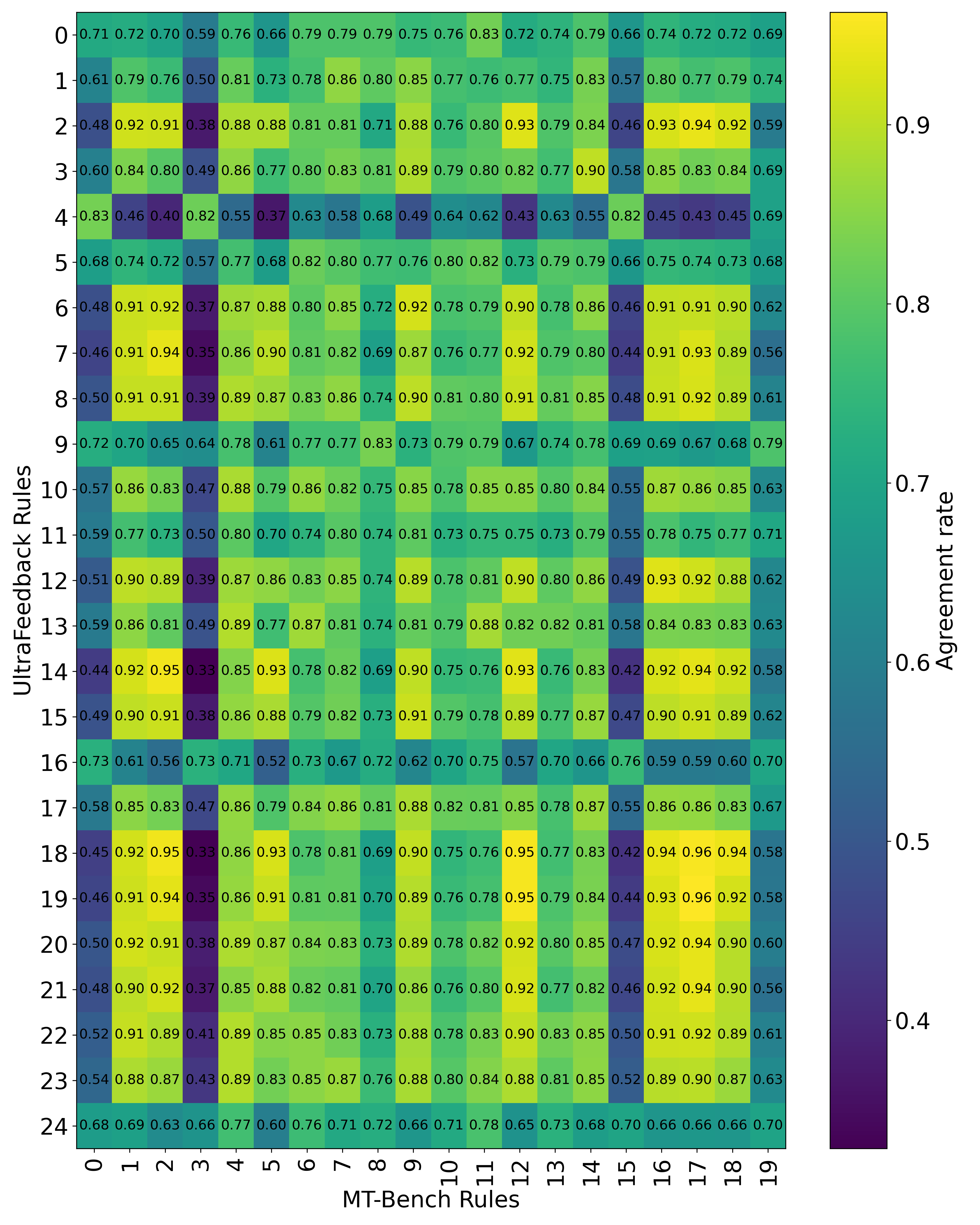}
  \caption{Rule agreement matrix on UltraFeedback data}
  \label{fig:rule-agreement-matrix-ultra}
\end{figure}

\begin{figure}[H]
  \centering
  \includegraphics[width=\textwidth]{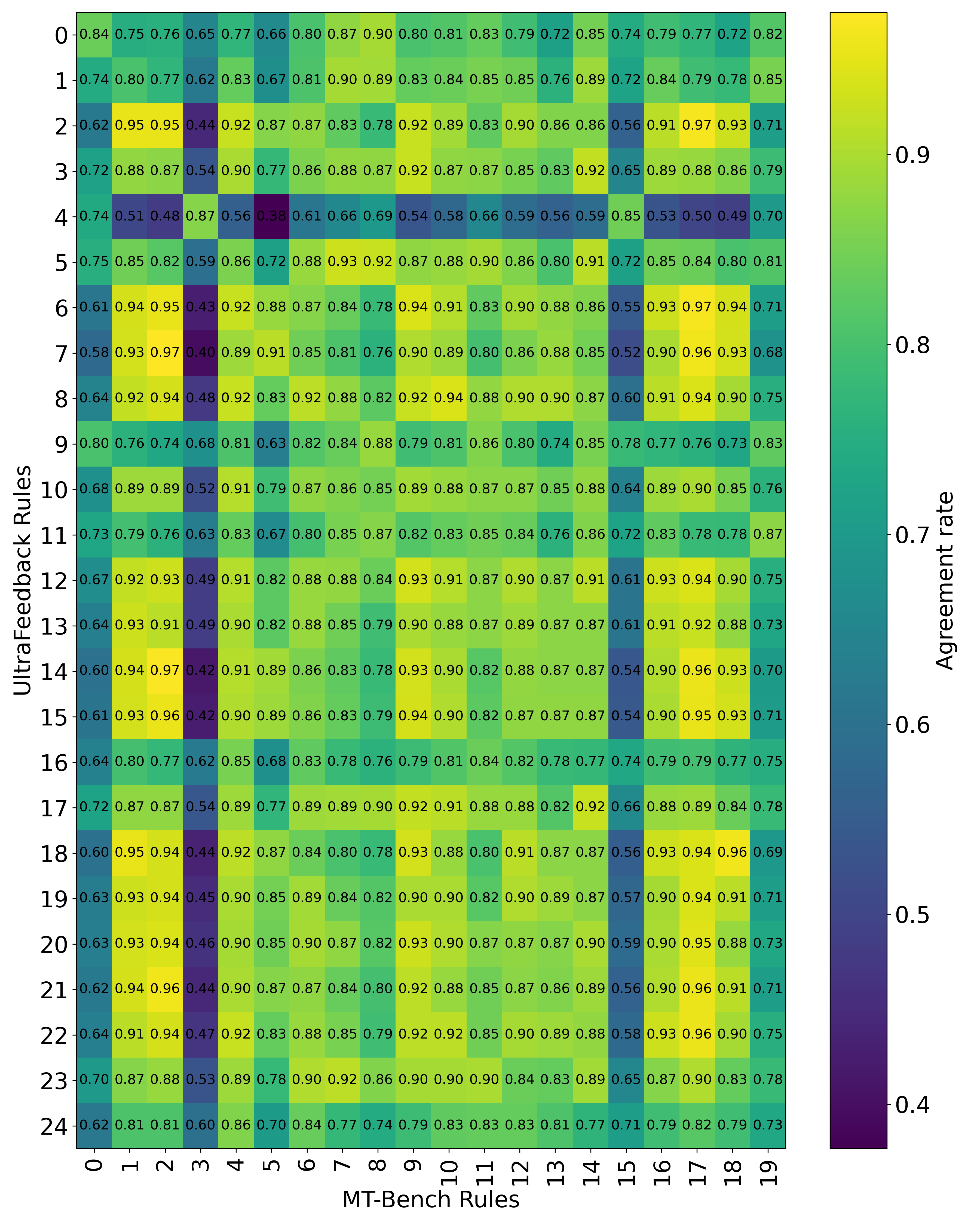}
  \caption{Rule agreement matrix on MT-Bench Human Judgements data}
  \label{fig:rule-agreement-matrix-mt}
\end{figure}

We showcase full rule agreement matrices between UltraFeedback and MT-Bench extracted rules on both UltraFeedback and MT-Bench data in Tables \ref{fig:rule-agreement-matrix-ultra} and \ref{fig:rule-agreement-matrix-mt}.

\section{Case study: Reasoning chain vs justification examples}
\label{sec:case-study}

\begin{figure}[htbp]
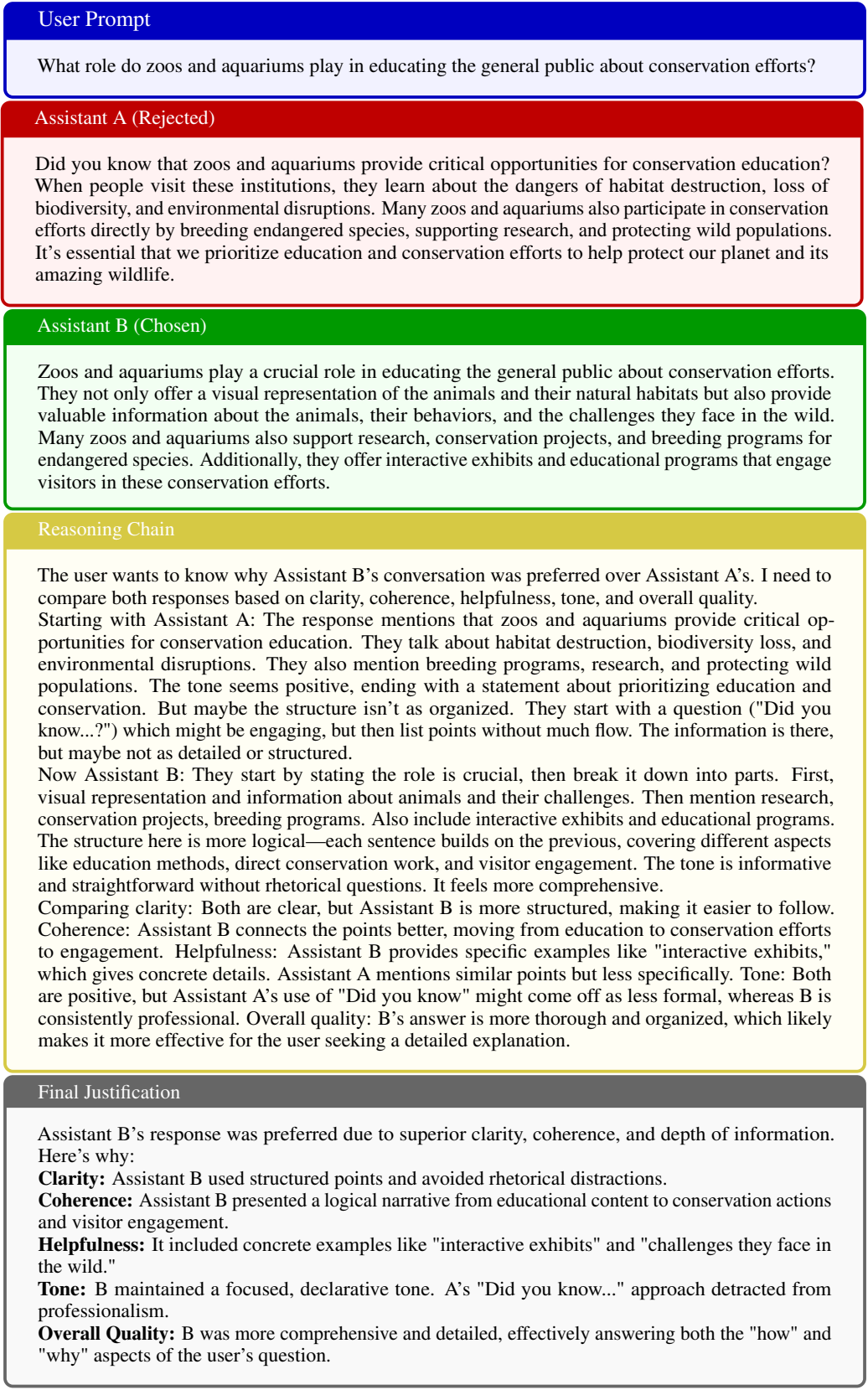

  \centering

  \begin{subfigure}[t]{0.98\textwidth}
    \begin{tcolorbox}[colback=blue!5!white, colframe=blue!75!black, title=User Prompt]
    \small
      What role do zoos and aquariums play in educating the general public about conservation efforts?
    \end{tcolorbox}
  \end{subfigure}

  \begin{subfigure}[t]{0.98\textwidth}
    \small
    \begin{tcolorbox}[colback=red!5!white, colframe=red!75!black, title=Assistant A (Rejected)]
      Did you know that zoos and aquariums provide critical opportunities for conservation education? When people visit these institutions, they learn about the dangers of habitat destruction, loss of biodiversity, and environmental disruptions. Many zoos and aquariums also participate in conservation efforts directly by breeding endangered species, supporting research, and protecting wild populations. It's essential that we prioritize education and conservation efforts to help protect our planet and its amazing wildlife.
    \end{tcolorbox}
  \end{subfigure} \hfill

  \begin{subfigure}[t]{0.98\textwidth}
    \small
    \begin{tcolorbox}[colback=green!5!white, colframe=green!60!black, title=Assistant B (Chosen)]
      Zoos and aquariums play a crucial role in educating the general public about conservation efforts. They not only offer a visual representation of the animals and their natural habitats but also provide valuable information about the animals, their behaviors, and the challenges they face in the wild. Many zoos and aquariums also support research, conservation projects, and breeding programs for endangered species. Additionally, they offer interactive exhibits and educational programs that engage visitors in these conservation efforts.
    \end{tcolorbox}
  \end{subfigure}

  \begin{subfigure}[t]{0.98\textwidth}
  \small
    \begin{tcolorbox}[colback=yellow!5!white, colframe=yellow!80!black, title=Reasoning Chain]
      The user wants to know why Assistant B's conversation was preferred over Assistant A's. I need to compare both responses based on clarity, coherence, helpfulness, tone, and overall quality.

      Starting with Assistant A: The response mentions that zoos and aquariums provide critical opportunities for conservation education. They talk about habitat destruction, biodiversity loss, and environmental disruptions. They also mention breeding programs, research, and protecting wild populations. The tone seems positive, ending with a statement about prioritizing education and conservation. But maybe the structure isn't as organized. They start with a question ("Did you know...?") which might be engaging, but then list points without much flow. The information is there, but maybe not as detailed or structured.

      Now Assistant B: They start by stating the role is crucial, then break it down into parts. First, visual representation and information about animals and their challenges. Then mention research, conservation projects, breeding programs. Also include interactive exhibits and educational programs. The structure here is more logical—each sentence builds on the previous, covering different aspects like education methods, direct conservation work, and visitor engagement. The tone is informative and straightforward without rhetorical questions. It feels more comprehensive.

      Comparing clarity: Both are clear, but Assistant B is more structured, making it easier to follow. Coherence: Assistant B connects the points better, moving from education to conservation efforts to engagement. Helpfulness: Assistant B provides specific examples like "interactive exhibits," which gives concrete details. Assistant A mentions similar points but less specifically. Tone: Both are positive, but Assistant A's use of "Did you know" might come off as less formal, whereas B is consistently professional. Overall quality: B's answer is more thorough and organized, which likely makes it more effective for the user seeking a detailed explanation.
    \end{tcolorbox}
  \end{subfigure}

  \begin{subfigure}[t]{0.98\textwidth}
    \small
    \begin{tcolorbox}[colback=gray!5!white, colframe=gray!80!black, title=Final Justification]
      Assistant B's response was preferred due to superior clarity, coherence, and depth of information. Here’s why:

      \textbf{Clarity:} Assistant B used structured points and avoided rhetorical distractions.

      \textbf{Coherence:} Assistant B presented a logical narrative from educational content to conservation actions and visitor engagement.

      \textbf{Helpfulness:} It included concrete examples like "interactive exhibits" and "challenges they face in the wild."

      \textbf{Tone:} B maintained a focused, declarative tone. A's "Did you know..." approach detracted from professionalism.

      \textbf{Overall Quality:} B was more comprehensive and detailed, effectively answering both the "how" and "why" aspects of the user’s question.
    \end{tcolorbox}
  \end{subfigure}

  \caption{Case study: reasoning chain vs justification.}
  \label{fig:case_study_conservation}
\end{figure}

To further analyze the importance of utilizing reasoning chains for rule extraction, we showcase a case study of a prompt asking about the role of zoos in animal conversation in Figure \ref{fig:case_study_conservation}. In the two model responses, Response A is clearly worse as it includes an out-of-place rhetorical question. While both the justification and the reasoning chain offer valid points, the reasoning chain is much more detailed and step-by-step in offering these points, including more examples and specific references to text, whereas the justification just provides a final summary.

By using a reasoning chain, our belief is that \method{} is able to extract better more in-depth and logical rules since the reasoning chain offers logical justifications rather than vague descriptions.

\newpage
\section{Prompts}
\label{sec:prompts}

We list the prompts used for the extraction process in Figures \ref{fig:prompt-justification}
, \ref{fig:prompt-rule-ext}, and \ref{fig:prompt-merge} respectively. Additionally, we include the prompts for rule verification in Figures \ref{fig:prompt-verify-concise} and \ref{fig:prompt-verify}, and the prompt used to determine UltraFeedback winner judgments for win-rate calculation in Figure \ref{fig:prompt-uf-judge}.

\begin{figure}[H]
  \centering
    \begin{tcolorbox}[colback=gray!5!white, colframe=gray!80!black, title=Justification Prompt]
[Instruction]\\
You are tasked with analyzing two conversations between an AI assistant and a user. Based on the content, please provide a detailed explanation of why the user might have preferred the winning conversation.\\
Please consider aspects such as clarity, coherence, helpfulness, tone, and overall quality.

[Conversation with Assistant A]

\{conversation\_a\}

[Conversation with Assistant B]

\{conversation\_b\}

[Winning Conversation]: \{winner\}

[Your Explanation]\\
    \end{tcolorbox}

  \caption{Justification (\method{} Extractor stage 1) prompt.}
  \label{fig:prompt-justification}
\end{figure}

\begin{figure}[H]
  \centering
    \begin{tcolorbox}[colback=gray!5!white, colframe=gray!80!black, title=Rule Extraction Prompt]
[Instruction]\\
Based on the following reasoning about why conversation with assistant {winner} is better, extract any rule-like statements implied by the reasoning that indicate this preference. Rule-like statements should be able to be judged objectively and deterministically. Below are a few examples of rule-like statements:\\
Example 1:\\
- The assistant's responses should validate any assumptions made with sufficient context and examples.\\
Example 2:\\
- The assistant's responses should not simply restate information provided by the user as its answer.\\
Example 3:\\
- The assistant's responses should have a structure that satisfies the user's request.\\
Return the list as a JSON array of strings. Do not use ```json```, just output the JSON array directly. If there are no rule-like statements, return an empty JSON array.

[Reasoning]\\
\{reasoning\_chain\}\\
    \end{tcolorbox}

  \caption{Rule extraction (\method{} Extractor stage 2) prompt.}
  \label{fig:prompt-rule-ext}
\end{figure}

\begin{figure}[H]
  \centering
    \begin{tcolorbox}[colback=gray!5!white, colframe=gray!80!black, title=Rule Merging Prompt]
[Instruction]\\
Below is a large list of rule-like statements regarding the behavior of an AI assistant. Some of these rules might be duplicates or very similar in meaning.\\
Please merge them so that there are no duplicates or rules with very similar meanings.\\
Return the merged list as a JSON array of strings. Do not use ```json```, just output the JSON array directly.

[Rules]\\
\{rules\_text\}\\

    \end{tcolorbox}

  \caption{Rule merging (\method{} Extractor stage 3) prompt.}
  \label{fig:prompt-merge}
\end{figure}

\begin{figure}[H]
  \centering
    \begin{tcolorbox}[colback=gray!5!white, colframe=gray!80!black, title=Rule Verifier Prompt]
You are an impartial judge. Determine whether the AI assistant's response in the following conversation both complies with the rule below and does so in a concise manner:\\
\\
Rule:\\
\{rule\}\\

[Start of Conversation]\\
\{conversation\}

[End of Conversation]\\

[Analysis]\\
Base your judgment solely on whether (1) the response satisfies the rule and (2) the response does so in a concise manner.\\
\\
Only respond with "[[Yes]]" if **both** conditions are fully satisfied. If either condition is not met, respond with "[[No]]". If the rule is not applicable to the task, treat it as satisfied.\\
\\
Respond with one of the following options, and nothing else: "[[Yes]]" or "[[No]]".\\
    \end{tcolorbox}

  \caption{Rule verifier prompt.}
  \label{fig:prompt-verify-concise}
\end{figure}

\begin{figure}[H]
  \centering
    \begin{tcolorbox}[colback=gray!5!white, colframe=gray!80!black, title=Rule Verifier Prompt (no conciseness)]
[Instruction]\\
Please act as an impartial judge and evaluate whether the responses provided by an AI assistant in the following conversation satisfy the following rule:\\
\{rule\}\\
Be as objective as possible when evaluating the rule and do not evaluate other characteristics of the response. If the rule is not applicable for this task, treat it as if the rule is satisfied. You must provide your answer by strictly outputting either one of the following two options: "[[Yes]]" or "[[No]]" and nothing else.

[Start of Conversation]\\
\{conversation\}

[End of Conversation]\\
    \end{tcolorbox}

  \caption{Rule verifier prompt (no conciseness).}
  \label{fig:prompt-verify}
\end{figure}

\begin{figure}[H]
  \centering
    \begin{tcolorbox}[colback=gray!5!white, colframe=gray!80!black, title=UF Win-rate Judgement Prompt]
I want you to create a leaderboard of different large-language models. To do so, I will give you the instructions (prompts) given to the models, and the responses of two models. Please rank the models based on which responses would be preferred by humans. All inputs and outputs should be python dictionaries.\\
\\
Here is the prompt:\\
\{\{\\
    "instruction": """\{instruction\}"""\\
\}\}\\
\\
Here are the outputs of the models:

\begin{tabbing}
[\\
\hspace{1em}\=\{\{\\
\hspace{2em}\=\="model": "model\_1",\\
\hspace{2em}\=\="answer": """\{output\_1\}"""\\
\hspace{1em}\=\}\},\\
\hspace{1em}\=\{\{\\
\hspace{2em}\=\="model": "model\_2",\\
\hspace{2em}\=\="answer": """\{output\_2\}"""\\
\hspace{1em}\=\}\}\\
]\\
\end{tabbing}

Now please rank the models by the quality of their answers, so that the model with rank 1 has the best output. Then return a list of the model names and ranks, i.e., produce the following output:

\begin{tabbing}
[\\
\hspace{1em}\=\{\{'model': \=<model-name>, 'rank': \=<model-rank>\}\},\\
\hspace{1em}\=\{\{'model': \=<model-name>, 'rank': \=<model-rank>\}\}\\
]\\
\end{tabbing}
Your response must be a valid Python dictionary and should contain nothing else because we will directly execute it in Python. Please provide the ranking that the majority of humans would give.

\quad
    \end{tcolorbox}

  \caption{UltraFeedback win-rate judgement prompt.}
  \label{fig:prompt-uf-judge}
\end{figure}

\newpage
\section{Compute Resources}
\label{sec:compute}

All training runs are conducted on a HPC cluster via the SLURM job management. All runs are conducted with 8 Nvidia L40S GPUs and 64 CPUs. The SFT and RM training runs have 256 GB CPU memory available, whereas the RL stage has 512 GB CPU memory available. We detail the time of execution for each of the training runs in Table \ref{tab:time}.

\begin{table}[H]
\centering
\small
\caption{Training run time rounded to the nearest hour.}
\label{tab:time}
\begin{tabular}{lccc}
\toprule
\textbf{Run} & \textbf{Time}\\
\midrule
SFT & 2 hrs \\
RM & 2 hrs \\ 
PPO & 2 hrs \\
GRPO (One episode) & 4 hrs \\
GRPO (Two episodes) & 13 hrs \\
GRPO + Length Control (One episode) & 3 hrs\\
GRPO + Length Control (Two episodes) & 12 hrs \\
GRPO + Length Penalty (One episode) & 3 hrs \\
GRPO + Length Penalty (Two episodes) & 10 hrs\\
\method{} (UF-extracted, one episode) & 15 hrs \\
\method{} (UF-extracted, two episodes) & 38 hrs \\
\method{} (UF justification-extracted) & 19 hrs \\
\method{} (UF-extracted) w/o Concise & 16 hrs \\
\method{} (UF-extracted) w/o Concise, Scaling & 15 hrs \\
\method{} (MT-extracted) & 14 hrs \\
\bottomrule
\end{tabular}
\end{table}

\section{Licenses}
\label{sec:licenses}

Asset URLS and licenses are displayed in Table \ref{tab:licenses}.

\begin{table}[H]
\centering
\small
\caption{Asset URLs and licenses. *Custom license available at \url{https://llama.meta.com/llama3/license}.}
\label{tab:licenses}
\begin{tabular}{p{2cm}p{5cm}p{33mm}l}
\toprule
\textbf{Asset} & \textbf{URL} & \textbf{Purpose} & \textbf{License}\\
\midrule
Llama-3-8B & \url{https://huggingface.co/meta-llama/Meta-Llama-3-8B} & Base model & Custom* \\
DeepSeek-R1 & \url{https://aws.amazon.com/bedrock/deepseek/} (Used on Bedrock) & Extraction process & MIT \\
UltraFeedback-Binarized & \url{https://huggingface.co/datasets/lmsys/mt_bench_human_judgments} & Dataset & MIT \\
MT-Bench Human Judgements & \url{https://huggingface.co/datasets/lmsys/mt_bench_human_judgments} & Dataset & CC-BY 4.0\\
LLM-as-a-judge code & \url{https://github.com/lm-sys/FastChat/tree/main/fastchat/llm_judge} & MT-Bench benchmark & Apache-2.0\\
AlpacaEval repo & \url{https://github.com/tatsu-lab/alpaca_eval} & AlpacaEval 2.0 benchmark & Apache-2.0 \\
PAR repo & \url{https://github.com/PorUna-byte/PAR} & Filtering code & MIT\\
OpenRLHF & \url{https://github.com/OpenRLHF/OpenRLHF} & Training framework & Apache-2.0\\
vLLM & \url{https://github.com/vllm-project/vllm} & Model inference & Apache-2.0\\
\bottomrule
\end{tabular}
\end{table}

\end{document}